\def\year{2021}\relax
\documentclass[letterpaper]{article} 

\usepackage{aaai21}

\usepackage[utf8]{inputenc} 
\usepackage{booktabs}       
\usepackage{amsfonts}       
\usepackage{nicefrac}       
\usepackage{microtype}      
\usepackage{times}  
\usepackage{helvet} 
\usepackage{courier}  
\usepackage[hyphens]{url}  
\usepackage{graphicx} 
\usepackage{natbib}  
\usepackage{caption} 
\usepackage{multirow}

\usepackage{bm}

\newcommand{\comment}[1]{}

\usepackage{graphicx}
\usepackage{subcaption} 
\usepackage[font=small,skip=0.3\baselineskip]{caption}
\usepackage{amsmath}
\usepackage{amssymb}
\usepackage[]{algorithm2e}

\usepackage{enumitem}

\newcommand{\norm}[1]{\left\|#1\right\|}
\newcommand{\Fb}{\mathbf{F}}
\newcommand{\Xb}{\mathbf{X}}
\newcommand{\Yb}{\mathbf{Y}}
\newcommand{\Wb}{\mathbf{W}}
\newcommand{\Ib}{\mathbf{I}}

\newcommand{\xb}{\mathbf{x}}
\newcommand{\yb}{\mathbf{y}}

\newcommand{\RR}{\mathbb{R}}
\newcommand{\NN}{\mathbb{N}}

\newcommand{\cN}{\mathcal{N}}
\newcommand{\cL}{\mathcal{L}}

\newcommand{\cB}{\mathcal{B}}

\title{Temporal Latent Auto-Encoder: A Method for Probabilistic Multivariate Time Series Forecasting
}

\author{%
  Nam Nguyen\footnote{Nam Nguyen and Brian Quanz contributed equally to this work}, Brian Quanz\footnotemark[1] \\}
\affiliations {
IBM Research\\
\{nnguyen,blquanz\}@us.ibm.com
}

\begin{document}

\maketitle

\begin{abstract}

Probabilistic forecasting of high dimensional multivariate time series is a notoriously challenging task, both in terms of computational burden and distribution modeling. Most previous work either makes simple distribution assumptions or abandons modeling cross-series correlations.  A promising line of work exploits scalable matrix factorization for latent-space forecasting, but is limited to linear embeddings, unable to model distributions, and not trainable end-to-end when using deep learning forecasting. We introduce a novel temporal latent auto-encoder method which enables nonlinear factorization of multivariate time series, learned end-to-end with a temporal deep learning latent space forecast model. By imposing a probabilistic latent space model, complex distributions of the input series are modeled via the decoder.
 Extensive experiments demonstrate that our model achieves state-of-the-art performance on many popular multivariate datasets, with gains sometimes as high as $50\%$ for several standard metrics.
\end{abstract}

\section{Introduction}
Forecasting - predicting future values of time series, is a key component in many industries \cite{fildes2008forecasting}.  Applications include forecasting supply chain and airline demand \cite{fildes2008forecasting,seeger2016bayesian}, financial prices \cite{kim2003financial}, and energy, traffic or weather patterns \cite{chatfield2000time}.  
Forecasts are often required for large numbers of related time series, i.e., multivariate time series forecasting, as opposed to univariate (single time series) forecasting.  For example, retailers may require sales/demand forecasts for millions of different products at thousands of different locations - amounting to billions of sales time series. 

In multivariate settings, one common approach is to fit a single multi-output model to predict all series simultaneously.  This includes statistical methods like vector auto-regressive (VAR) models \cite{lutkepohl2005new} and generalizations (e.g., MGARCH \cite{bauwens2006multivariate}), and multivariate state-space models \cite{durbin2012time}, as well as deep neural net (DNN) models including recurrent neural networks (RNNs) \cite{funahashi1993approximation}, temporal convolutional neural networks (TCNs) \cite{bai2018empirical}, and combinations \cite{lai2018modeling,goel2017r2n2,borovykh2017conditional,cheng2019towards,dasgupta2017nonlinear,cirstea2018correlated,rodrigues2020beyond}.  However, these are prone to overfitting and not scalable as the number of time series increases \cite{yu2016trmf,sen2019deepglo,salinas2018copula}.  

As such, another popular approach is to abandon multivariate forecasting entirely and perform univariate forecasting (i.e., fit a separate model per series). Classical statistical forecasting methods using simple parametric models of past values and forecasts are still arguably most commonly used in industry, such as auto-regressive AR and ARIMA models \cite{hyndman2018forecasting}, exponential smoothing (ES) \cite{mckenzie1984general}, and more general state-space models \cite{hyndman2008forecasting}.  
Such methods have consistently out-performed machine learning methods such as RNNs in large scale forecasting competitions until recently \cite{makridakis2020forecasting,makridakis2018statistical,makridakis2020m4,crone2011advances,benidis2020neural}. 
A key reason for recent success of deep learning for forecasting is multi-task univariate forecasting - sharing deep learning model parameters across all series, possibly with some series-specific scaling factors or parametric model components \cite{salinas2019deepar,smyl2020hybrid,bandara2020,li2019transformer,wen2017multi,rangapuram2018dsp,chen2018tada}.  E.g., the winner of the M4 forecasting competition \cite{makridakis2020m4} was a hybrid ES-RNN model \cite{smyl2020hybrid}, in which a single shared univariate RNN model is used to forecast each series but seasonal and level ES parameters are simultaneously learned per series to normalize them.  

However, a fundamental limitation of multi-task univariate forecasting approaches is they are unable to model cross-series correlations/effects \cite{rangapuram2018dsp,salinas2018copula}, common in many domains \cite{salinas2018copula,tsay2013multivariate,rasul2020normflow}.  For example, in retail, cross-product effects (e.g., increased sales of one product causing increased/decreased sales of related products) are well known \cite{gelper2016,leeflang2008,srinivasan2005identifying}.  In  financial time series one stock price may depend on relative prices of other stocks; and energy time series may have spatial correlations and dependencies.  Furthermore, these approaches cannot leverage the extra information provided from related series in case of noise or sparsity.  E.g., sales are often sparse (e.g., one sale a month for a particular product and store), so the sales rate cannot be accurately estimated from a single series.

A  promising line of research we focus on that addresses limitations of both the single, large multi-output multivariate model and the multi-task  univariate model approaches is to use factorization \cite{yu2016trmf,sen2019deepglo}.   Relationships between time series are factorized into a low rank matrix, i.e., each time series is modeled as a linear combination of a smaller set of latent, basis (or \emph{global}) time series, so forecasting can be performed in the low-dimensional latent space then mapped back to the input (\emph{local}) space. Thus modeling can scale to very large number of series while still capturing cross-series relationships.  Temporal regularized matrix factorization (TRMF) \cite{yu2016trmf} imposes temporal regularization on the latent time series so they are predictable by linear auto-regressive models.  Recently, DeepGLO \cite{sen2019deepglo} extended this approach to enable non-linear latent space forecast models.  DeepGLO iteratively alternates between linear matrix factorization and fitting a latent space TCN; forecasts from this model are then fed as covariates to a separately trained multi-task univariate TCN model.

However, these have several key limitations.  First, they cannot capture nonlinear relationships between series via the transformation, which are common in many domains.  
E.g., products' sales or stocks' prices may depend on relative price compared to others  (i.e., value ratios, a non-linear relationship).  Second, although deepGLO introduces deep learning, it is not an end-to-end model.  Since factorization is done separately and heuristic, alternating optimization with no convergence guarantees is used, the process is inefficient and may not find an optimal solution.  Third, they have no way to provide probabilistic outputs (i.e., predictive distributions), which are critical for practical use of forecasts \cite{makridakis2020forecasting}.  Fourth, they are limited to capturing stationary relationships between time series with the fixed linear transform on a single time point - whereas relationships between series are likely often nonstationary.

To address these limitations and extend the factorization line of research, we propose the Temporal Latent Autoencoder (TLAE) (see Figure \ref{fig:tlae}), which enables non-linear transforms of the input series trained end-to-end with a DNN temporal latent model to enforce predictable latent temporal patterns, and implicitly infers the joint predictive distribution simultaneously. Our \textbf{main contributions} are:  
\begin{itemize}[noitemsep,leftmargin=*]
  \item We enable nonlinear factorization for the latent temporal factorization line of research; we generalize the linear mappings to/from the latent space to nonlinear transforms by replacing them with encoder and decoder neural networks, with an input-output reproduction objective, i.e., an autoencoder \cite{kramer1991nonlinear,hinton1994autoencoders}. 
  Further, the autoencoder can use temporal models (e.g., RNNs) - so embeddings can evolve over time (be nonstationary). 
  
  \item We introduce temporal regularization in the latent space with flexible deep learning temporal models that can be trained end-to-end with stochastic gradient descent, by combining the objectives for reconstruction and forecast error in the latent and input spaces in the loss function.
  
  \item We enable probabilistic output sampling by injecting noise in the latent space prior to latent forecast decoding, so the model learns to implicitly model the cross-time-series joint predictive distribution by transforming the noise, similar to variational autoencoders (VAEs) \cite{doersch2016tutorial,kingma2014}.  Unlike VAEs, the latent mean (output of the latent forecast model) is not constrained.
  
  \item We perform extensive experiments with multiple  multivariate forecasting datasets, demonstrating superior performance compared to past global factorization approaches as well as comparable or superior performance to other recent state of the art forecast methods, for both point and probabilistic predictions (Section \ref{sec:experiments}).  We also provide a variety of analyses including hyper parameter sensitivity. 
\end{itemize}



\section{Related Work}
\label{sec:related}
Neural nets have a long history of applications in forecasting \cite{zhang1998forecasting,benidis2020neural}, historically mostly focused on univariate models.  Here we discuss details of recent related deep learning work beyond those mentioned in the introduction. For further details on classical methods please refer to \cite{hyndman2018forecasting,lutkepohl2005new,durbin2012time,bauwens2006multivariate}. We compare with representative classical forecast methods in experiments - e.g., VAR, ARIMA, and state space models (ETS). 

A trend in DNN forecasting is to normalize series to address different scaling / temporal patterns \cite{lai2018modeling,zhang2003time,salinas2019deepar,goel2017r2n2,cheng2019towards,bandara2020,smyl2020hybrid}.  E.g., LSTNet \cite{lai2018modeling} fits the sum of a linear AR model and a DNN with convolutional and recurrent layers.  A popular multi-task univariate forecast method, DeepAR \cite{salinas2019deepar}, scales each series by its average and fits a shared RNN across series.  Another recent sate-of-the-art multi-task univariate model \cite{li2019transformer} combines TCN embeddings with the Transformer architecture \cite{vaswani2017attention}.  
Although these work well on some datasets, as mentioned they are limited in use as they cannot model dependencies between series.

TADA\cite{chen2018tada}, DA-RNN \cite{qin2017dual} and GeoMAN \cite{liang2018geoman} use encoder-decoder approaches built on sequence-to-sequence work \cite{cho2014learning,bahdanau2015neural}.  However the encoder-decoder is not an autoencoder, is designed for factoring in exogenous variables for multi-step univariate forecasting - not modeling cross series relationships / multivariate forecasting, and is not probabilistic.
An autoencoder was used in \cite{cirstea2018correlated}, but only for pre-processing / denoising of individual series before training an RNN, so did not consider factorizing cross-series relationships or deriving probabilistic outcomes, as in our method. 

Recently a few DNN models have also been proposed to model multivariate forecast distributions \cite{salinas2018copula,wang2019deep,rasul2020normflow}.  A low-rank Gaussian copula model was proposed \cite{salinas2018copula} in which a multitask univariate LSTM \cite{hochreiter1997long} is used to output transformed time series and diagonal and low-rank factors of a Gaussian covariance matrix.  However, it is limited in flexibility / distributions it can model, sensitive to choice of rank, and difficult to scale to very high dimensional settings.  
A deep factor generative model was proposed \cite{wang2019deep} in which a linear combination of RNN latent global factors plus parametric noise models the local series distributions.
However, this can only model linear combinations of global series and specific noise distributions, has no easy way to map from local to global series, 
and is inefficient for inference and learning (limited network and data size that can be practically used).  
Further, a recent concurrent work uses normalizing flows for probabilistic forecasting \cite{rasul2020normflow}: a multivariate RNN is used to model the series progressions (single large multi-output model), with the state translated to the output joint distribution via a normalizing flow approach  \cite{dinh2017density}.
However, invertible flow requires the same number of latent dimensions as input dimensions, so it does not scale to large numbers of time series. E.g., the temporal model it is applied across all series instead of a low dimensional space as in our model, so for RNN it has quadratic complexity in the number of series, whereas ours can be much lower (shown in supplement).



Another line of related work is on variational methods with sequence models such as variational RNN (VRNN) \cite{chung2015recurrent} and \cite{chatzis2017recurrent}, e.g., VRNN applies a VAE to each hidden state of an RNN over the input series.  Both of these apply the RNN over the input space so lack scalability benefits and require propagating multi-step predictions through the whole model, unlike our method which scalably applies the RNN and its propagation in a low-dimensional latent space.  Further, due to noise added at every time step, VRNN may struggle with long term dependencies, and the authors state the model is designed for cases of high signal-to-noise ratio, whereas most forecast data is very noisy.

\section{Problem Setup and Methodology}

{\bf Notation. } A matrix of multivariate time series is denoted by a bold capital letter, univariate series by bold lowercase letters. Given a vector $\xb$, its $i$-th element is denoted by $x_i$. For a matrix $\Xb$, we use $\xb_i$ as the $i$-th column and $x_{i,j}$ is the $(i,j)$-th entry of $\Xb$. $\norm{\Xb}_{\ell_2}$ is the matrix Frobenius norm. $\norm{\xb}_{\ell_p}$ is the $\ell_p$-norm of the vector $\xb$, defined as $(\sum_i x_i^p)^{1/p}$. Given a matrix $\Yb \in \RR^{n \times T}$, $\Yb_B$ is indicated as a sub-matrix of $\Yb$ with column indices in $B$. For a set $\cB$, $|\cB|$ is regarded as the cardinality of this set. Lastly, for functions $f$ and $g$, $f \circ g$ is the composite function, $f \circ g(\xb) = f(g(\xb))$.  

{\bf Problem definition.} Let a collection of high dimensional multivariate time series be denoted by $( \yb_1,..., \yb_T) $, where each $\yb_i$ at time point $i$ is a vector of dimension $n$. Here we assume $n$ is often a large number, e.g., ${\sim}10^3$ to $10^6$ or more.
We consider the problem of forecasting $\tau$ future values $(\yb_{T+1}, ..., \yb_{T+\tau})$ of the series given its observed history $\{ \yb_i \}_{i=1}^T$. A more difficult but interesting problem is modeling the conditional probability distribution of the high dimensional vectors:
\begin{equation}
\textstyle
    p(\yb_{T+1}, ..., \yb_{T+\tau} | \yb_{1:T} ) = \prod_{i=1}^{\tau} p(\yb_{T+i} | \yb_{1:T+i-1} ).
\end{equation}
This decomposition turns the problem of probabilistically forecasting several steps ahead to rolling prediction: the prediction at time $i$ is input to the model to predict the value at time $i+1$. 
Next we describe our key contribution ideas in deterministic settings, then extend it to probabilistic ones.

\subsection{Point prediction}
\label{subsec:point_prediction}

{\bf Motivation. } Temporal regularized matrix factorization (TRMF) \cite{yu2016trmf}, decomposes the multivariate time series represented as a matrix $\Yb \in \RR^{n \times T}$ (composed of $n$ time series in its rows) into components $\Fb \in \RR^{n \times d}$ and $\Xb \in \RR^{d \times T}$ while also imposing temporal constraints on $\Xb$. The matrix $\Xb$ is expected to inherit temporal structures such as smoothness and seasonality of the original series. If $\Yb$ can be reliably represented by just the few time series in $\Xb$, then tasks on the high-dimensional series $\Yb$ can be performed on the much smaller dimensional series $\Xb$. 
In \cite{yu2016trmf} forecasting future values of $\Yb$ is replaced with the much simpler task of predicting future values on the latent series $\Xb$, so the $\Yb$ prediction is just a weighted combination of the new $\Xb$ values with weights defined by the matrix $\Fb$.

To train temporal DNN models like RNNs, data is batched temporally. 
Denote $\Yb_B$ as a batch of data containing a subset of $b$ time samples, $\Yb_B = [\yb_t, \yb_{t+1},..., \yb_{t+b-1}]$ where $B = \{t,...,t+b-1\}$ are time indices. To perform constrained factorization \cite{yu2016trmf} proposed to solve:
\begin{equation}
\min_{ \Xb, \Fb, \Wb } \cL(\Xb, \Fb, \Wb) = \frac{1}{|\cB|} \sum_{B \in \cB} \cL_B (\Xb_B, \Fb, \Wb),
\end{equation}
where $\cB$ is the set of all data batches and each batch loss is:
\begin{equation}
 \textstyle
 \cL_B(\Xb_B, \Fb, \Wb) \triangleq  \frac{1}{nb} \norm{\Yb_B - \Fb \Xb_B}_{\ell_2}^2 + \lambda \mathcal{R} (\Xb_B; \Wb).
\end{equation}
Here, $\mathcal{R}(\Xb_B; \Wb)$ is regularization parameterized by $\Wb$ on $\Xb_B$ to enforce certain properties of the latent factors and $\lambda$ is the regularization parameter. In order to impose temporal constraints, \cite{yu2016trmf} assumes an autoregressive model on $\Xb_B$ specified simply as $\xb_\ell = \sum_{j = 1}^L \Wb^{(j)} \xb_{\ell - j} $ where $L$ is a predefined lag parameter. Then, the regularization reads
\begin{equation}
\label{eq:autoregressive regularization}
\mathcal{R} (\Xb_B; \Wb) \triangleq \sum_{\ell=L+1}^b \norm{ \xb_\ell - \sum_{j = 1}^L \Wb^{(j)} \xb_{\ell - j} }_{\ell_2}^2.
\end{equation}
The optimization is solved via alternating minimization with respect to variables $\Xb, \Fb$, and $\Wb$.

Recently, \cite{sen2019deepglo} considered applying deep learning to the same problem; the authors proposed to replace the autoregressive component with a temporal convolutional network (TCN) \cite{bai2018empirical}. Their TCN-MF model employed the following regularization 
\begin{equation}
\label{eq:tcn regularization}
\mathcal{R} (\Xb_B; \Wb) \triangleq \sum_{\ell=L+1}^b \norm{ \xb_\ell -  \text{TCN} (\xb_{\ell-L, ..., \xb_{\ell-1}}; \Wb) }_{\ell_2}^2,
\end{equation}
where $\Wb$ is the set of parameters of the TCN network; alternating minimization was also performed for optimization.

\cite{sen2019deepglo} also investigated feeding TCN-MF predictions as ``global'' features into a ``local'' multi-task model forecasting individual time series. However, as mentioned, both \cite{yu2016trmf} and \cite{sen2019deepglo} have several challenging limitations. First, due to the linear nature of the matrix factorization, the models implicitly assume linear relationships across time series. This implies the models will fail to capture non-linear correlation cross series (e.g., one series inversely proportional to another) that often occurs in practice, separately from the global temporal patterns. 
Second, implementation of these optimization problems with alternating minimization is sufficiently involved, especially when the loss has coupling terms as in (\ref{eq:autoregressive regularization}). In \cite{sen2019deepglo}, although the simple autoregressive model is replaced by a TCN, this network cannot incorporate the factorization part, making back-propagation impossible to perform end-to-end. TCN-MF model is therefore unable to leverage recent deep learning optimization developments. This may explain why solutions of TCN-MF are sometimes sub-optimal as compared to the simpler TRMF approach \cite{yu2016trmf}.

{\bf Our model.} In this paper we propose a new model to overcome these weaknesses. We  observe that if $\Yb$ can be decomposed exactly by $\Fb$ and $\Xb$, then $\Xb = \Fb^+ \Yb$ where $\Fb^+$ is the pseudo-inverse of $\Fb$. This implies that
$\Yb = \Fb \Fb^+ \Yb$.

Now if $\Fb^+$ can be replaced by an encoder and $\Fb$ by a decoder, we can exploit the ideas of autoencoders \cite{kramer1991nonlinear,hinton1994autoencoders} to seek more powerful nonlinear decompositions. The latent representation is now a nonlinear transformation of the input, $\Xb = g_{\bm \phi}(\Yb)$ where $g_{\bm \phi}$ is the encoder that maps $\Yb$ to $d$ dimensional $\Xb$: $g: \RR^n \rightarrow \RR^d$ and $\bm \phi$ is the set of parameters of the encoder. The nonlinearity of the encoder allows the model to represent more complex structure of the data in the latent embedding. The reconstruction of $\Yb$ is $\hat{\Yb} = f_{\bm \theta}(\Xb)$ where $f_{\bm \theta}$ is the decoder that maps $\Xb$ back to the original domain: $f: \RR^{d} \rightarrow \RR^n$ and $\bm \theta$ is the set of parameters associated with the decoder.

\begin{figure*}[h!]
\centering
    \includegraphics[width=140mm]{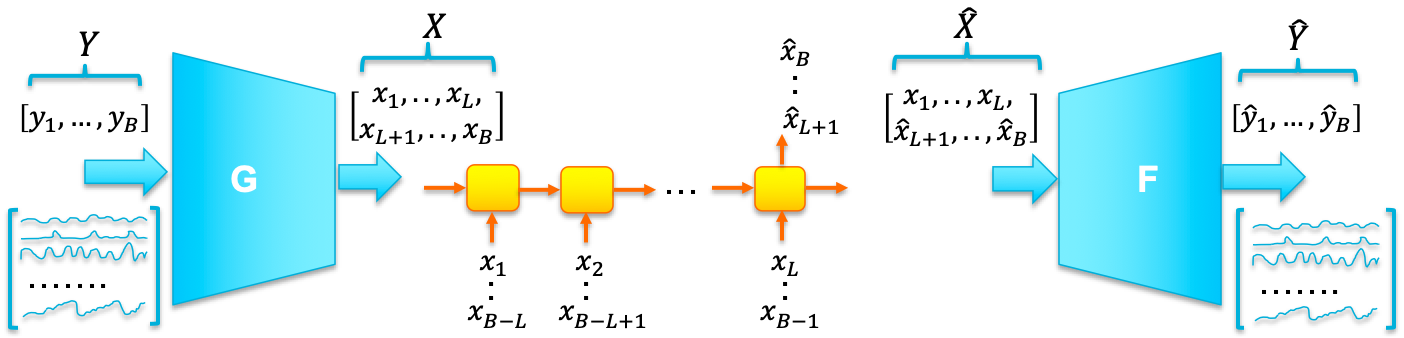}
    \caption{Temporal latent autoencoder.  Though illustrated with an RNN, any temporal DNN model (e.g., TCN or Transformer) can be used in the latent space. The decoder translates Normal noise to arbitrary distributions.}
  \label{fig:tlae} 
\end{figure*}

Additionally, we introduce a new layer between the encoder and decoder to capture temporal structure of the latent representation. The main idea is illustrated in Figure \ref{fig:tlae}; in the middle layer an LSTM network \cite{hochreiter1997long} is employed to encode the long-range dependency of the latent variables. The flow of the model is as follows: a batch of the time series $\Yb_B = [\yb_1, ..., \yb_b] \in \RR^{n \times b}$ is embedded into the latent variables $\Xb_B = [\xb_1, ..., \xb_b] \in \RR^{d \times b}$ with $d \ll n$. These sequential ordered $\xb_i$ are input to the LSTM to produce outputs $\hat{\xb}_{L+1}, ..., \hat{\xb}_b$ with each $\hat{\xb}_{i+1} = h_{\Wb} (\xb_{i-L+1},...,\xb_{i})$ where $h$ is the mapping function. $h$ is characterized by the LSTM network with parameters $\Wb$. The decoder will take the matrix $\hat{\Xb}_B$ consisting of variables $\xb_1, ..., \xb_{L}$ and $\hat{\xb}_{L+1}, ..., \hat{\xb}_b$ as input and yield the matrix $\hat{\Yb}_B$. 

As seen from the figure, batch output $\hat{\Yb_B}$ contains two components. The first, $\hat{\yb}_i$ with $i=1,...,L$, is directly transferred from the encoder without passing through the middle layer: $\hat{\yb}_i = f_{\bm \theta} \circ g_{\bm \phi} (\yb_i)$, while the second component $\hat{\yb}_i$ with $i=L+1,...,b$ is a function of the past input: $\hat{\yb}_{i+1} = f_{\bm \theta} \circ h_{\Wb} \circ g_{\bm \phi} (\yb_{i-L+1}, ..., \yb_{i})$. By minimizing the error $\|\hat{\yb}_i - \yb_i\|^p_{\ell_p}$, one can think of this second component as providing the model the capability to predict the future from the observed history, while at the same time the first one requires the model to reconstruct the data faithfully.

The objective function with respect to a batch of data is defined as follows.
\begin{equation}
\label{eq:tlae batch}
\begin{split}
\textstyle
&\cL_B (\Wb, \bm \phi, \bm \theta) \triangleq \frac{1}{nb} \norm{\Yb_B - \hat{\Yb}_B}^p_{\ell_p} \\
&+ \lambda \frac{1}{d(b-L)} \sum_{i=L}^{b-1} \norm{\xb_{i+1} - h_{\Wb}(\xb_{i-L+1},...,\xb_{i})}^q_{\ell_q},
\end{split}
\end{equation}
and the overall loss function is 
\begin{equation}
\label{eq:tlae}
\textstyle
\min_{ \Wb, \bm \phi, \bm \theta } \cL(\Wb, \bm \phi, \bm \theta ) = \frac{1}{|\cB|} \sum_{B \in \cB} \cL_B (\Wb, \bm \phi, \bm \theta).
\end{equation}
On the one hand, by optimizing $\hat{\Yb}$ to be close to $\Yb$, the model is expected to capture the correlation cross time series and encode this global information into a few latent variables $\Xb$. On the other hand, minimizing the discrepancy between $\hat{\Xb}$ and $\Xb$ allows the model to capture temporal dependency and provide the predictive capability of the latent representation.   We add a few more remarks:
\begin{itemize}[noitemsep,leftmargin=*]
    \item Although we use LSTMs here, other networks (e.g., TCN \cite{bai2018empirical} or Transformer \cite{vaswani2017attention}) can be applied. 
    \item A fundamental difference between TRMF / TCN-MF and our method is that in the former, latent variables are part of the optimization and solved for explicitly while in ours, latent variables are parameterized by the networks, thus back-propagation can be executed end-to-end for training.
    \item By simpler optimization, our model allows more flexible selection of loss types imposed on $\hat{\Yb}$ and $\hat{\Xb}$. In experiments, we found that imposing $\ell_1$ loss on $\hat{\Yb}$ consistently led to better prediction while performance remains similar with either $\ell_1$ or $\ell_2$ loss on $\hat{\Xb}$. Since  $\ell_1$ loss is known to be more robust to outliers, imposing it directly on $\hat{\Yb}$ makes the model more resilient to potential outliers.
    \item Encoders/decoders themselves can use temporal DNNs so non-static relationships can be captured.
\end{itemize}

Once the model is learned, forecasting several steps ahead is performed via rolling windows. Given past input data $[\yb_{T-L+1},...,\yb_T]$, the learned model produces the latent prediction $\hat{\xb}_{T+1} = h_{\Wb} (\xb_{T-L+1},..., \xb_T)$ where each $\xb_i = g_{\bm \phi} (\yb_i)$. The predicted $\hat{\yb}_{T+1}$ is then decoded from $\hat{\xb}_{T+1}$. The same procedure can be sequentially repeated $\tau$ times (in the latent space) to forecast $\tau$ future values of $\Yb$ in which the latent prediction $\hat{\xb}_{T+2}$ utilizes $[\xb_{T-L+1},...,\xb_T, \hat{\xb}_{T+1}]$ as the input to the model. Notice that the model does not require retraining during prediction as opposed to TRMF.

\subsection{Probabilistic prediction}
\label{subsec:prob_prediction}

One of the notorious challenges with high-dimensional time series forecasting is how to probabilistically model the future values conditioned on the observed sequence: $p(\yb_{T+1},..., \yb_{T+\tau} | \yb_1,...,\yb_{T})$. Most previous works either focus on modelling each individual time series or parameterizing the conditional probability of the high dimensional series by a multivariate Gaussian distribution. However, this is inconvenient since the number of learned parameters (covariance matrix) grows quadratically with the data dimension. Recent DNN approaches make distribution assumptions (such as low-rank covariance) that limit flexibility and/or similarly lack scalability (see Section \ref{sec:related}).


In this paper, instead of directly modelling in the input space, we propose to encode the high dimensional data to a much lower dimensional embedding, on which a probabilistic model can be imposed. Prediction samples are later obtained by sampling from the latent distribution and translating these samples through the decoder. If the encoder is sufficiently complex so that it can capture non-linear correlation among series, we can introduce fairly simple probabilistic structure on the latent variables and are still able to model complex distributions of the multivariate data via the decoder mapping. Indeed, together with the proposed network architecture in Figure \ref{fig:tlae}, we model
\begin{equation}
\label{eq:xi distribution}
\textstyle
p(\xb_{i+1} | \xb_1, ..., \xb_i) =  \cN(\xb_{i+1} ; \bm{\mu}_i, \bm{\sigma}_i^2) , \quad i=L,..., b.
\end{equation}
Here, we fix the conditional distribution of latent variables to multivariate Gaussian with diagonal covariance matrix with variance $\bm{\sigma}_i^2$. This is meaningful as it encourages the latent variables to capture different orthogonal patterns of the data, which makes the representation more powerful, universal, and interpretable. The mean $\bm{\mu}_i$ and variance $\bm{\sigma}_i^2$ are functions of $\xb_{1},...,\xb_i$: $\bm{\mu}_i = h^{(1)}_{\Wb} (\xb_{1},...,\xb_i)$ and $\bm{\sigma}_i^2 = h^{(2)}_{\Wb} (\xb_{1},...,\xb_i)$.


The objective function $\cL_B (\bm \phi, \theta, \Wb)$ with respect to the batch data $\Yb_B$ is defined as the weighted combination of the reconstruction loss and the negative log likelihood loss
\begin{equation}
\label{eq:batch loss of prob tlae}
\textstyle
    \frac{1}{nb} \norm{\hat{\Yb}_B - \Yb_B}_{\ell_p}^p - \lambda \frac{1}{b-L} \sum_{i=L+1}^{b} \log \cN(\xb_{i} ; \bm{\mu}_{i-1}, \bm{\sigma}_{i-1}^2).
\end{equation}

This bears similarity to the loss of the variational autoencoder (VAE) \cite{kingma2014autoencoder} which consists of a data reconstruction loss and a Kullback–Leibler divergence loss encouraging the latent distribution to be close to the standard multivariate Gaussian with zero mean and unit diagonal covariance.  Unlike VAEs, our model has a temporal model in the latent space and is measuring a conditional discrepancy (with no fixed mean constraint). Further, rather than encourage unit variance we fix latent space unit variance, to also help avoid overfitting during training - i.e., we set $\bm{\sigma}_i^2=1$ in our model. As with GANs, the decoder learns to translate this  noise to arbitrary distributions (examples in supplement).

Recall the output of the decoder $\hat{\yb}_{i+1} = f_{\bm \theta} (\hat{\xb}_{i+1})$ for each sample $\hat{\xb}_{i+1}$ from the distribution (\ref{eq:xi distribution}). In order to back-propagate through batch data, we utilize the reparameterization trick as in VAEs. I.e., $\hat{\xb}_{i+1} = {\bm \mu_i} + {\bm 1} \epsilon = h^{(1)}_{\Wb} (\xb_{1},...,\xb_i) + {\bm 1} \epsilon$ with $\epsilon \sim \cN(0,1)$. Each iteration when gradient calculation is required, a sample $\epsilon$ is generated which yields latent sample $\hat{\xb}_{i+1}$ and associated $\hat{\yb}_{i+1}$. 

Once the model is learned, next prediction samples $\hat{\yb}_{T+1}$ can be decoded from samples $\hat{\xb}_{T+1}$ of the latent distribution (\ref{eq:xi distribution}). Conditioned on $\xb_{T-L+1},..., \xb_T$ and the mean of $\hat{\xb}_{T+1}$, samples $\hat{\xb}_{T+2}$ can also be drawn from (\ref{eq:xi distribution}) to obtain prediction samples $\hat{\yb}_{T+2}$, and so on.



\section{Experiments}
\label{sec:experiments}



\subsection{Point estimation}
\label{sec:point_estimate_exp}
We first evaluate our point prediction model with loss defined in  (\ref{eq:tlae}). We compare with state-of-the art multivariate and univariate forecast methods \cite{sen2019deepglo} \cite{yu2016trmf} \cite{salinas2019deepar} using 3 popular datasets: \emph{traffic}: hourly traffic of 963 San Fancisco car lanes \cite{cuturi2011fast,dua2019uci}, \emph{electricity}: hourly consumption of 370 houses \cite{electricityUCI}, and \emph{wiki}: daily views of \textasciitilde 115k Wikipedia articles \cite{wikiKaggle}. 
Traffic and electricity show weekly cross-series patterns; wiki contains a very large number of series. 
Following conventional setups \cite{salinas2019deepar,sen2019deepglo,yu2016trmf}, we perform rolling prediction evaluation: 24 time-points per window, last 7 windows for testing for traffic and electricity, and 14 per window with last 4 windows for wiki. We use the last few windows prior to the test period for any hyper parameter selection. We use 3 standard metrics: mean absolute percent error (MAPE), weighted absolute percent error (WAPE), and symmetric MAPE (SMAPE) to measure test prediction error. Dataset / formula details are in the supplement. 

Network architecture and optimization setup in experiments is as follows: the encoder and decoder use feed forward network (FNN) layers with ReLU activations on all but the last layer. Layer dimensions vary per dataset. The network architecture in the latent space is a 4-layer LSTM, each with 32 hidden units. In all experiments, $\ell_1$ loss is used on $\hat{\Yb}$ and $\ell_2$ for the regularization. Regularization parameter $\lambda$ is set to $0.5$. We find the $\ell_1$ loss on $\hat{\Yb}$ can help reduce potential outlier effects and provide more stable and accurate results. 
Setup and training details are provided in the supplement.

Table \ref{compare with deeglo} shows the comparison of different approaches. All results except our proposed TLAE were reported in \cite{sen2019deepglo} under the same experimental setup; we pick the best reported results in \cite{sen2019deepglo} with or without data normalization. Here, global models use global features for multivariate forecasting while local models employ univariate models and separately predict individual series. Here we do not compare our model with conventional methods (e.g., VAR, ARIMA) since they are already confirmed to obtain inferior performance to TRMF and DeepAR methods \cite{yu2016trmf} \cite{salinas2019deepar}.  


As seen in the table, our method significantly out-performs other global modeling / factorization methods on all datasets (8/9 dataset-metric combinations) - showing it clearly advances the state-of-the-art for global factorization multivariate forecasting approaches.   Compared with other global models, the gain on traffic and electricity datasets can be as significant as $50\%$. Further, even without any local modeling and additional exogenous features like hour of day (as used in local and combined methods), our  method still achieves superior performance on 2/3 datasets across all metrics. 
Our model could likely be further improved by incorporating the exogenous features in the latent space or with local modeling (as done with deepGLO) - the point is our model provides a better global fit starting point. Also note our model only applied standard network architectures and did not make use of recent advanced ones such as TCNs or Transformer, for which we might expect further improvement.

Furthermore, in experiments latent dimensions are set to $16$ for traffic and $32$ for electricity data, as opposed to $64$ dimensions used in \cite{sen2019deepglo}. This indicates our model is able to learn a better and more compact representation. We show examples of learned latent series and input and latent space predictions (and distributions) in the supplement, illustrating our model is able to capture shared global patterns. We also highlight that our model does not need retraining during testing.

\begin{table*}[h!]
\centering
\caption{Comparison of different algorithms with WAPE/MAPE/SMAPE metrics. Only local and combined models employ additional features such as hour of day and day of week. Best results for global modeling methods are labeled in bold, best overall with $^*$. Our scores are the average over $3$ separate runs with different random initializations. Standard dev. is less than $0.003$ for all the metrics.}
\label{compare with deeglo}
\scriptsize
\begin{tabular}{  l | l | l | l | l }
Model & Algorithm & 
\multicolumn{3}{ |c }{Datasets} \\
& &Traffic & Electricity-Large & Wiki-Large \\ 
\hline \hline
\multirow{4}{*}{Global factorization}
 & \textbf{TLAE} (our proposed method) & $\bf 0.117^* / 0.137^* / 0.108^*$ & ${\bf 0.080^*}/{\bf 0.152^*}/ {\bf 0.120^*}$ & $0.334 / {\bf 0.447} / {\bf 0.434}$ \\ 
 & DeepGLO - TCN-MF \cite{sen2019deepglo} & $0.226/0.284/0.247$ & $0.106/0.525/0.188 $ &  $0.433/1.59/0.686$ \\
 & TRMF \cite{yu2016trmf} (retrained) & $0.159/0.226/ 0.181$ & $0.104/0.280/0.151$ & ${\bf 0.309}/0.847/0.451$  \\
 & SVD+TCN & $0.329/0.687/0.340$ & $0.219/0.437/0.238$ & $0.639/2.000/0.893$ \\
\hline
\multirow{5}{*}{Local \& combined} & DeepGLO - combined \cite{sen2019deepglo}  & $0.148/0.168/0.142$ & $   0.082 /0.341/  0.121$ & $ 0.237/0.441/0.395$ \\
 & LSTM & $0.270/0.357/0.263$ & $0.109/0.264/0.154$ & $0.789/0.686/0.493$ \\
 & DeepAR \cite{salinas2019deepar} & $0.140/0.201/ 0.114$ & $0.086/0.259/ 0.141$ & $0.429/2.980/0.424$ \\
 & TCN (no LeveledInit) & $0.204/0.284/0.236$ & $0.147/0.476/0.156$ & $0.511/0.884/0.509$ \\
 & TCN (LeveledInit) & $0.157/0.201/0.156$ & $0.092/0.237/0.126$ & $ 0.212^*/0.316^*/0.296^*$ \\
 & Prophet \cite{taylor2018forecasting} & $0.303/0.559/0.403$ & $0.197/0.393/0.221$ & - \\
 \hline
\end{tabular}
\end{table*}

\subsection{Probabilistic estimation}

Our next experiments consider probabilistic forecasting. We compare our model with the state-of-the-art probabilistic multivariate method \cite{salinas2018copula}, as well as \cite{wang2019deep} and univariate forecasting \cite{salinas2019deepar,rangapuram2018dsp,li2019transformer} in the supplement, each following the same data setup (note: different data splits and processing than in Section \ref{sec:point_estimate_exp}; details in supplement). We apply the same network architecture as in previous experiments, except the latent variable loss is the negative Gaussian log likelihood (\ref{eq:batch loss of prob tlae}) and the regularization parameter $\lambda$ is set to $0.005$. A smaller $\lambda$ is selected in this case to account for the scale difference between the regularizations in (\ref{eq:tlae batch}) and (\ref{eq:batch loss of prob tlae}).
Latent samples are generated during training with the reparameterization trick and distribution statistics obtained from decoded sampled latent predictions. 
Two additional datasets are included: \emph{solar} (hourly production from 137 stations) and \emph{taxi} (rides taken at 1214 locations every 30 minutes) (training / data details in the supplement. 

For probabilistic estimates, we report both the continuous ranked probability score across summed time series (CRPS-sum) \cite{matheson1976scoring,gneiting2007strictly,salinas2018copula} (details in supplement) and MSE (mean square error) error metrics, to measure overall joint distribution pattern fit and fit of joint distribution central tendency, respectively, so that together the two metrics give a good idea of how good the predictive distribution fit is.
Results are shown in Table \ref{table:CRPS-sum/MSE} comparing error scores of TLAE with other methods reported in \cite{salinas2018copula}. 
Here, GP is the Gaussian process model of \cite{salinas2018copula}. As one can observe, our model outperforms other methods on most of the dataset-metric combinations (7/10), in which the performance gain is significant on Solar, Traffic, and Taxi datasets. We also provide additional tables in the supplement to show CRPS and MSE scores with standard deviation from different runs for more thorough comparison. In the supplement, we visually show different latent series learned from the model on all datasets as well as predictive distributions and sampled 2D joint distributions, demonstrating non-Gaussian and non-stationary distribution patterns. From the plots we see that some are focused on capturing global, more slowly changing patterns across time series; others appear to capture local, faster changing information. Combinations of these enable the model to provide faithful predictions.




\begin{table*}[h!]
\centering
\caption{Comparison of different algorithms with CRPS-Sum and MSE metrics. Most results are from Tables 2 and 6 of \cite{salinas2018copula} with VRNN and our results (under same setup) at the end (TLAE). Lower scores indicate better performance. Shows the mean score from $3$ separate runs with random initialization. VAR and GARCH are traditional statistical multivariate methods \cite{lutkepohl2005new,bauwens2006multivariate}; Vec-LSTM methods use a single global LSTM that takes and predicts all series at once, with different output Gaussian distribution approaches; and GP methods are DNN gaussian process ones proposed in \cite{salinas2018copula} with GP-Copula the main one - see details in \cite{salinas2018copula}. A '-' indicates a method failed (e.g., required too much memory as not scalable enough for data size).}
\label{table:CRPS-sum/MSE}
\scriptsize
\begin{tabular}{  l  l  l  l l l }
& CRPS-Sum / MSE & & & &\\
\hline
Estimator & Solar & Electricity-Small & Traffic & Taxi & Wiki-Small \\
\hline
VAR & 0.524 / 7.0e3 & 0.031 / 1.2e7 & 0.144 / 5.1e-3 & 0.292 / - & 3.400 / - \\
GARCH & 0.869 / 3.5e3 & 0.278 / 1.2e6 & 0.368 / 3.3e-3 & - / - & - / - \\
Vec-LSTM-ind & 0.470 / 9.9e2 & 0.731 / 2.6e7 & 0.110 / 6.5e-4 & 0.429 / 5.2e1 & 0.801 / 5.2e7 \\
Vec-LSTM-ind-scaling & 0.391 / 9.3e2 & 0.025 / 2.1e5 & 0.087 / 6.3e-4 & 0.506 / 7.3e1 & 0.113 / 7.2e7 \\
Vec-LSTM-fullrank & 0.956 / 3.8e3 & 0.999 / 2.7e7 & -/- & -/- & -/- \\
Vec-LSTM-fullrank-scaling & 0.920 /3.8e3 & 0.747 / 3.2e7 & -/- & -/- & -/- \\
Vec-LSTM-lowrank-Copula & 0.319 / 2.9e3 & 0.064 / 5.5e6 & 0.103 / 1.5e-3 & 0.4326 / 5.1e1 & 0.241 / 3.8e7 \\
LSTM-GP \cite{salinas2018copula} & 0.828 / 3.7e3 & 0.947 / 2.7e7 & 2.198 / 5.1e-1 & 0.425 / 5.9e1 & 0.933 / 5.4e7 \\
LSTM-GP-scaling \cite{salinas2018copula} & 0.368 / 1.1e3 & {\bf 0.022}  / \bf 1.8e5 & 0.079 / 5.2e-4 & 0.183 / 2.7e1 & 1.483 / 5.5e7  \\
LSTM-GP-Copula \cite{salinas2018copula} & 0.337 / 9.8e2 &  0.024 / 2.4e5 & 0.078 / 6.9e-4 & 0.208 / 3.1e1 & {\bf 0.086} / 4.0e7 \\
VRNN \cite{chung2015recurrent} & 0.133 / 7.3e2 &   0.051 / 2.7e5 & 0.181 / 8.7e-4 & 0.139 / 3.0e1 & 0.396 / 4.5e7 \\
{\bf TLAE} (our proposed method) &  \bf 0.124 / 6.8e2  & 0.040 / 2.0e5 & {\bf 0.069 / 4.4e-4 } & {\bf 0.130 / 2.6e1} & 0.241 / \bf 3.8e7 
\end{tabular}
\end{table*}

\subsection{Hyper parameter sensitivity \& ablation study}
\label{sec:param_sensitivity}
We conducted various experiments with traffic data to monitor the prediction performance of the model when varying different hyper parameters: batch size, regularization parameter $\lambda$, and latent dimension. We use the same network architecture as the previous section and train the model with probabilistic loss in (\ref{eq:batch loss of prob tlae}). Predictions are obtained by decoding the mean of the latent distribution and the prediction accuracy is measured by MAPE, WAPE, and SMAPE metrics.  

As explained in our proposed model in Figure \ref{fig:tlae}, the latent sequence to the decoder consists of two sub-sequences $\{\xb_1,...,\xb_{L}\}$ and $\{\hat{\xb}_{L+1},...,\hat{\xb}_{b} \}$. While the first one is directly transmitted from the encoder, the second one is the output of the LSTM network. Minimizing the discrepancy between $\hat{\xb}_{L+i}$ and $\xb_{L+i}$ equips the latent variables with the predictive capability, which implicitly contributes to better prediction of the time series, so we would expect selecting the batch size sufficiently larger than $L$ (the number of LSTM time steps) should lead to better predictive performance.

\begin{figure}[!htb]
\minipage{0.22\textwidth}
  \includegraphics[width=\linewidth]{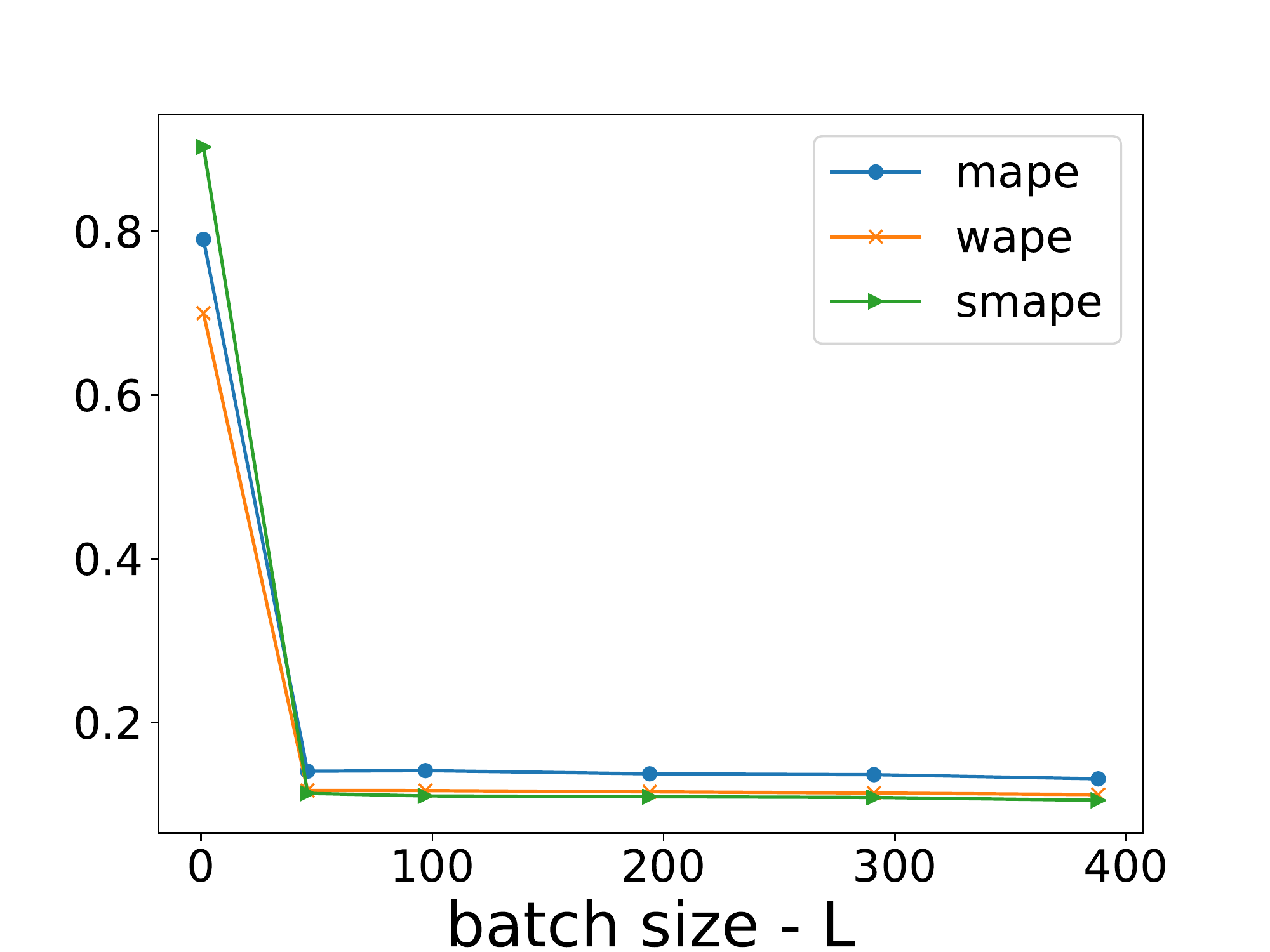}
  \caption{Varying batch size}\label{fig:vary bs}
\endminipage\hfill
\minipage{0.22\textwidth}
  \includegraphics[width=\linewidth]{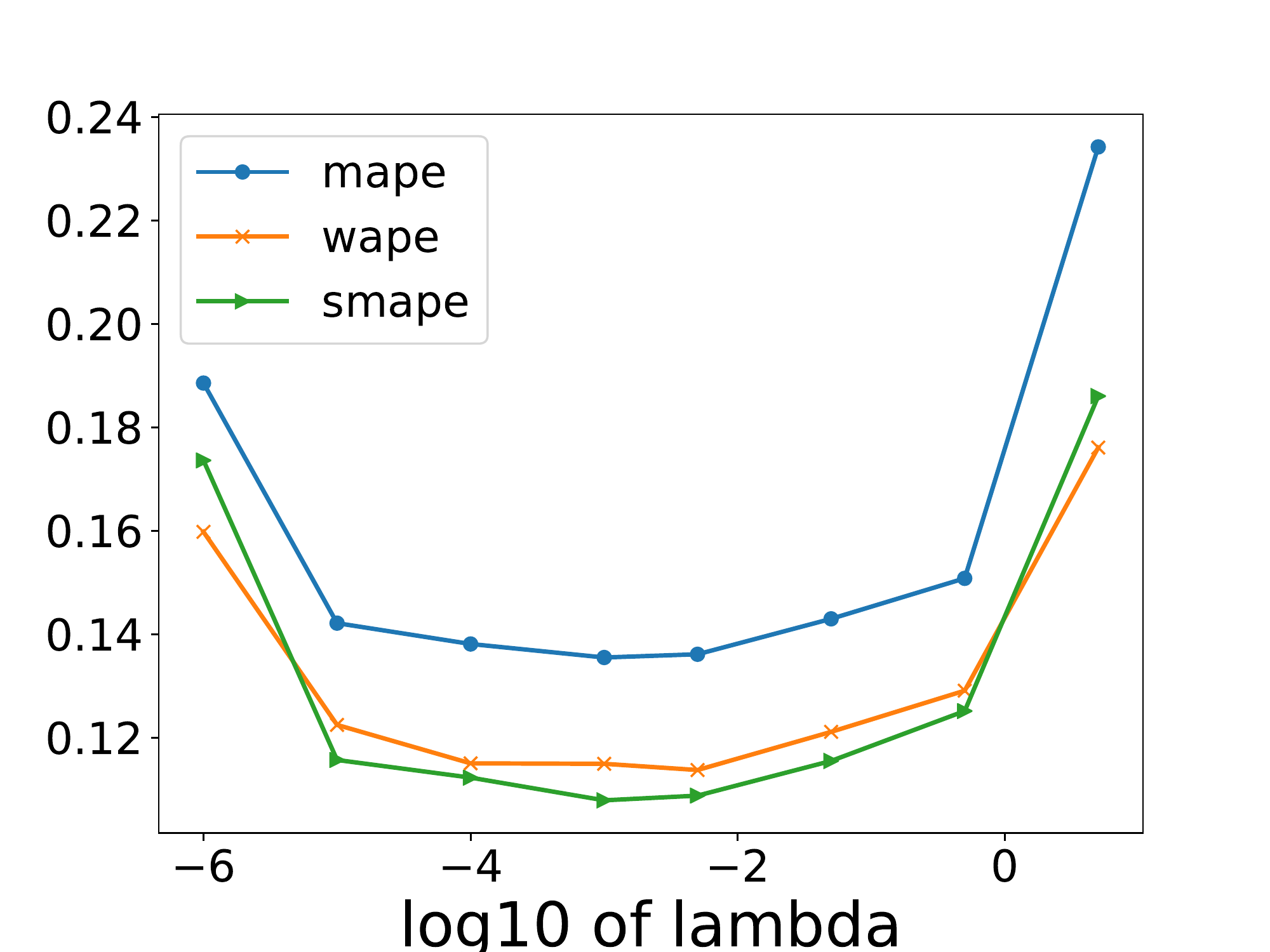}
  \caption{Varying $\lambda$}\label{fig:vary lambda}
\endminipage\hfill
\minipage{0.22\textwidth}%
  \includegraphics[width=\linewidth]{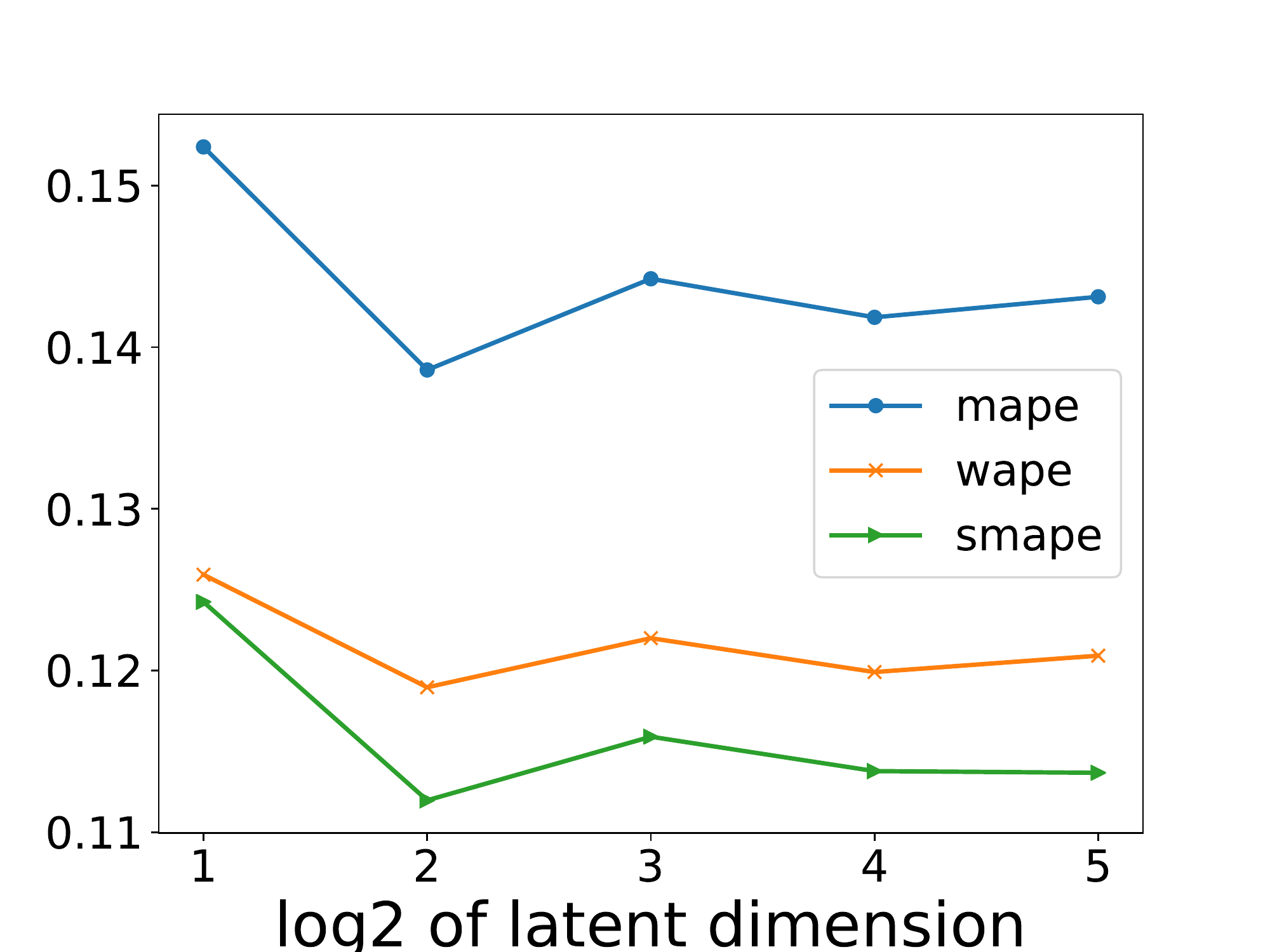}
  \caption{Varying latent dimension}\label{fig:vary latent dim}
\endminipage
\end{figure}

We validate this intuition by optimizing the model with varying batch size $b = L+1, 1.5L, 2L, 2.5L$, and $3L$ where $L$ is set to $194$. Figure \ref{fig:vary bs} illustrates the variability of the prediction accuracy with increasing batch size. As one can observe, all three metrics decreases as we increase the batch size, confirming the importance of balancing the two latent sub-sequences, i.e., having a balance between both a direct reproduction and predictive loss component in the input space.

Next, for fixed batch size $b = 2L$, we vary $\lambda$ = 1e-6, 1e-5, 1e-4, 1e-3, 5e-3, 5e-2, 5e-1, and $5$ and train the model with $500$ epochs. Figure \ref{fig:vary lambda} plots the prediction accuracy with respect to different choices of $\lambda$. As the regularization on the latent space is ignored by assigning very small parameter $\lambda$, the overall prediction performance is poor. The performance is quickly improved when higher $\lambda$ is selected - the latent constraint starts to dominate the reconstruction term with larger $\lambda$. The best range of $\lambda$ is between [1e-4, 1e-2].


Lastly, we vary the latent embedding dimension between $[2,4,8,16,32]$, and train with $200$ epochs vs. $1000$ to reduce computational time. Figure \ref{fig:vary latent dim} shows the impact on metrics. The model performance slightly improves with increase of the latent dimension and starts to stabilize, indicating  higher latent dimension may not help much.


Additionally, to validate the hypothesis that nonlinear transformation helps, we performed ablation study by using a linear decoder and encoder under the same setup.  We found worse performance than the non-linear case, though still better than DeepGLO (details in supplement).


\section{Conclusion}

This paper introduces an effective method for high dimensional multivariate time series forecasting, advancing the state-of-the-art for global factorization approaches. The method offers an efficient combination between flexible non-linear autoencoder mapping and inherent latent temporal dynamics modeled by an LSTM. The proposed formulation allows end-to-end training and, by modelling the distribution in the latent embedding, generating complex predictive distributions via the non-linear decoder.  Our experiments illustrate the superior performance compared to other state-of-the-art methods on several common time series datasets. 
Future directions include testing temporal models in the autoencoder, $3D$ tensor inputs, and combinations with local modeling. 

\clearpage
{\small
\bibliography{reference}}

\clearpage

\section{Supplementary Material}
The following sections are included in this supplementary materials document:
\begin{itemize}
    \item \textbf{Section \ref{sub:dataset_details} - Dataset details}: 
    More detailed  description of the data sets used and their setups, including table listing the attributes of the data.
    \item \textbf{Section \ref{sub:netarch_and_learning} - Network architecture and learning:}
    More detailed description of the network architectures used and selected via validation set tuning per dataset, and the training and hyper parameter selection process.
    \item \textbf{Section \ref{sub:computational_complexity} -  Model complexity and run times:} 
    Analysis of the computational complexity, showing the benefit of our method as compare to input-space modeling, for scaling to large number of time series, as well as recorded run times for training on the different datasets.
    \item \textbf{Section \ref{sub:ablation} - Benefit of nonlinear factorization - Ablation Study:}
    Ablation study results in which we compare forecasting performance with using just a linear encoder vs. the nonlinear encoder of TLAE - demonstrating the benefit of introducing nonlinear embedding.
    \item \textbf{Section \ref{sub:additional_comparisons} - Additional comparisons, metrics, and result details:}
    Results for additional comparisons for additional metrics including CRPS (non-sum version), and including the standard deviation across multiple random runs.
    \item \textbf{Section \ref{sec:univariate_comparisons} -  Comparison to univariate methods}: Here we give detailed restults on our experiment comparing TLAE to recent state-of-the-art univariate forecasting methods.
    \item \textbf{Section \ref{sub:visualization} -  Visualization - of losses, predictions, and predictive distributions}: Plots of the different train loss components and test losses per epoch, as well as plots of predictions in the latent and input spaces, including distribution plots (showing prediction intervals from sampled probabilistic predictions illustrating heteroskedastic predictive distributions).  Additionally, $2$D scatter plots are provided showing examples of the learned joint predictive distribution amongst pairs of time series are shown, illustrating the non-Gaussian and sometimes multi-modal and nonlinear relationships between series, which our model is able to capture.
    \item \textbf{Section \ref{sub:tensor_extension} -  Extension to tensor data for incorporating exogenous variables}: Details on the extension of our method to include additional exogenous features in our model using tensor inputs.
    \item \textbf{Section \ref{subsec:evaluation metrics} -  Evaluation metrics}: Complete details and formulas for the different evaluation metrics used in the experiments.
\end{itemize}


\subsection{Dataset details}
\label{sub:dataset_details}

\begin{table*}[htb!]
\begin{center}
\caption{Dataset statistics for both deterministic and probabilistic forecast experiments. Electricity (large), and wiki (large) datasets are used for deterministic forecasting, while smaller datasets are used for probabilistic forecasting (following previous state-of-the-art forecast method setups used for comparison \cite{salinas2018copula}).}
\label{table:dataset statistics for probabilistic forecast}
\scriptsize
\begin{tabular}{  l | l  l  l  l  l  l}
Dataset & time steps $T$ & dimension $n$ & $\tau$ (number predicted steps) & k (rolling windows) & frequency & domain \\
\hline
Traffic & $10392$ & $963$ & $24$ & $7$ & hourly & $\RR^+$ \\
Electricity (large) & $25920$ & $370$ & $24$ & $7$ & hourly & $\RR^+$ \\
Electricity (small) & $5833$ & $370$ & $24$ & $7$ & hourly & $\RR^+$ \\
Solar & $7009$ & $137$ & $24$ & $7$ & hourly & $\RR^+$  \\
Taxi & $1488$ & $1214$ & $24$ &  $56$ & $30$-minutes & $\NN$  \\
Wiki (large) & $635$ & $115084$ & $14$ & $4$ & daily & $\NN$ \\
Wiki (small) & $792$ & $2000$ & $30$ & $5$ & daily & $\NN$ \\
\end{tabular}
\end{center}
\end{table*}

Table \ref{table:dataset statistics for probabilistic forecast} describes statistics of datasets used in our experiments. Only the traffic data is the same on both experiments, the electricity and wiki datasets utilize different portions of the data. In particular, electricity (large) and wiki (large) are used for deterministic experiments while the smaller ones are for probabilistic experiments (taking the same setup as previous state-of-the-art probabilistic methods to enable comparison).  Note we used the same setup in each case so we could directly compare with the best results prior state-of-the-art methods were able to achieve.  For our point forecast we compared with the state-of-the-art global factorization approaches designed to be scalable, which is why they were applied on larger data previously.  For the probabilistic case we did not find other work applying to larger data, due to lack of scalability these others typically can only be applied on the smaller versions of the data - highlighting another key benefit of our work, as it provides a highly scalable probabilistic forecasting method (as none of the prior global factorization methods could provide probabilistic forecasts). 

In these tables, $T$ is the total number of time steps in each dataset for training and $n$ is the number of dimensions (number of series). We perform rolling prediction for the test evaluation, where $\tau$ is the number of predicted steps at each window and $k$ is the number of rolling windows. In particular, in the first window, we use $ \yb_{T-L},..., \yb_{T}$ as the input of size $L$ ($L$ is the LSTM time steps) to predict the next $\tau$ steps: $\hat{\yb}_{T+1},..., \hat{\yb}_{T+\tau}$. The next window, the actual input $ \yb_{T-b+ \tau},..., \yb_{T+\tau}$ is considered to predict $\hat{\yb}_{T+\tau+1},..., \hat{\yb}_{T+2\tau}$, and so on until $k$ rolling windows is reached. The prediction $\hat{\yb}_{T},..., \hat{\yb}_{T+k\tau}$ and their actual numbers are used to calculate the score metrics. The model training data is $\yb_1,..., \yb_{T}$ and model is not allowed to see test window samples during training or when predicting those test windows. The datasets are publicly available in the github repositories of \cite{sen2019deepglo} and \cite{salinas2018copula} - in the included program code supplement we provide instructions and links for downloading each dataset used.  Descriptions of all 5 datasets are as follows:
\begin{itemize}
    \item \emph{traffic}: Hourly traffic (in terms of occurpancy rates between 0 and 1) of 963 San Fancisco freeway car lanes, collected across 15 months \cite{cuturi2011fast,dua2019uci}
    \item \emph{electricity}: Hourly electricity consumption of 370 households/clients, collected over 4 years \cite{electricityUCI}. The full dataset is used for deterministic forecast while a smaller subset of data is used for probablistic forecast \cite{salinas2018copula}. 
    \item \emph{wiki}: Daily page views of \textasciitilde 115k Wikipedia articles over a couple years \cite{wikiKaggle} (note, a smaller subset of 2000 selected cleaner article series was used for probabilistic evaluation, following \cite{salinas2018copula}). 
    \item \emph{solar}\footnote{https://www.nrel.gov/grid/solar-power-data.html}: The solar power production records from January to August in 2006, which is sampled every hour from 137 PV plants in Alabama \cite{lai2018modeling}.
    \item \emph{taxi}\footnote{https://www1.nyc.gov/ site/tlc/about/tlc-trip-record-data}: Half hourly traffic of New York taxi rides taken at $1214$ locations during the month of January $2015$ for training and January $2016$ for testing. 
\end{itemize}

\subsection{Network architectures and learning}
\label{sub:netarch_and_learning}

\begin{table*}[h!]
\begin{center}
\caption{Network architectures for different datasets.}
\label{table:architectures}
\scriptsize
\begin{tabular}{  l | l  l  l  l  l  l}
Dataset  & Encoder type & Encoder and Decoder & LSTM layers & LSTM hidden dim & sequence length & number epochs \\
\hline
Traffic & FFN & $[96, 64, 32, 16]$ & $4$ & $32$ & $194$ & $1000$ \\
Electricity (large) & FFN & $[96, 64, 32]$ & $4$ & $64$ & $194$ & $750$ \\
Electricity (small) & FFN & $[128, 64]$ & $4$ & $64$ & $194$ & $200$ \\
Solar & FFN & $[64, 16]$ & $4$ & $32$ & $194$ & $300$ \\
Taxi & FFN & $[64, 16]$ & $4$ & $32$ & $120$ & $100$ \\
Wiki (large) & FFN & $[64, 32]$ & $4$ & $32$ & $128$ & $20$ \\
Wiki (small) & FFN & $[64, 32]$ & $4$ & $32$ & $128$ & $20$ \\
\end{tabular}
\end{center}
\end{table*}

For all the experimental setup, we design the encoder and decoder to contain a few feed forward layers, each followed by a ReLU activation, except the last layer, and an LSTM network to model the temporal dependency in the latent space. Details of the network architectures of datasets used in our paper are provided in Table \ref{table:architectures} (note the decoder sequence of layer sizes is reversed from that of the encoder).  The process of hyper parameter selection using a held-out valid set is described later - but only minor hyper-parameter tuning was used to select some modeling settings (such as trying a few different number of layers / neurons). To train the model, we use the Adam optimizer with the learning rate set to $0.0001$ (which is commonly used). We generally found that larger learning rates lead to unstable performance. The batch size is set to be twice the number of LSTM timesteps. This is to have an even balance between the number of latent features directly transferring to the decoder and the number passing through the middle LSTM layer before the decoder (please see the hyper-parameter sensitivity results in the main paper to see the effect of varying this). The running time is dependent on the size of the datasets and the number of epochs. For instance, with traffic training data consisting of $963$ series, each has more than $10,344$ time samples, it takes roughly 4 days to train with $1000$ epochs with 4 CPUs. Detail of the running time is provided in Table \ref{table:running time}. Recently after performing the experiments we have also started testing running using GPUs and found significant speed up - taking hours instead of days in most cases, but have not performed full analysis with the GPU case.

We experimented with sacrificing reasonable run time for better accuracy by trying to train the traffic data with $2000$ epochs.  We found this gave sufficiently more accurate performance than the result reported in Table 1. In particular, WAPE/MAPE/SMAPE with $2000$ epochs is $0.106/0.118/0.092$.  This suggests with even more training epochs we might have the chance to improve our results reported throughout the paper further for some data sets - because we did not thoroughly search all hyper-parameters for our method during hyper parameter selection with the validation set, e.g., the number of epochs, the best possible results for our method may not have been realized and shown here. Further, we believe the running time can be sufficiently reduced if GPUs are employed, as mentioned.

Note also in Table \ref{table:architectures} the relatively smaller number of epochs used in the case of Wiki.  For this dataset we generally found the training would converge after much fewer epochs, perhaps because of the much smaller number of time points, and the larger decrease in size transforming to the latent space.  It is future work to try more hyper parameter settings for the Wiki data as well.

Across all the experiments, we select the regularization parameter $\lambda=0.5$ for the deterministic model and $\lambda=0.005$ when working with probabilistic model. These choices are to guarantee the reconstruction and regularization losses in the objective have similar order of magnitude (note: please see the hyper parameter sensitivity study section in the main paper to see impact of changing $\lambda$). The loss function with respect to $\Yb$ is chosen to be the $\ell_1$ norm. This choice is particularly natural to handle possible outliers in the dataset. We also tested the model with the $\ell_2$ loss but the outcome is generally not as good as the $\ell_1$ counterpart. For the regularization loss in the latent space, we apply $\ell_2$ loss for the deterministic model while negative log likelihood is applied for probabilistic model, as described in the main text. Our experiments also indicate that similar performance is attained with either $\ell_2$ or $\ell_1$ imposed on the regularization. To select batch data for training, one can consider taking sliding windows where the two batches are entirely overlapping except one time point. For example, one can take two batches $\Yb_{\tau:\tau+b}$ and $\Yb_{\tau+1 : \tau+b+1}$ at times $\tau$ and $\tau+1$ where $b$ is the batch size and feed to the model. However, to speed up the training process, we select the non-overlapping region to be $24$, $12$, $12$, $2$, and $1$ for traffic, electricity, solar, taxi, and wiki datasets, respectively. I.e., smaller non-overlapping window size was used with smaller datasets.

In our experiments, we split the total training data $\Yb = \{ \yb_1,..., \yb_{T}\}$ into two sets, the first set $\{\yb_1,...,\yb_{T-m} \}$ is for modeling training and the last set $\{\yb_{T-m},...,\yb_T \}$ for validation where $m$ is set to the size of the test data. We tested the model with a few different architectural choices by changing the FFN layers between $[96,64,32,16]$, $[96,64,32]$, $[64, 32]$, $[64, 16]$, and $[128, 64]$, changing the LSTM hidden units between $32$ and $64$, and with different number of epochs. The best network architecture on the validation set was retrained with full training $\{\yb_1,...,\yb_T \}$ data and used for testing. We did not perform a thorough architecture and hyperparameter search. Doing so may help improve performance further, as suggested by our hyper-parameter sensitivity analyses in the Hyper Parameter Sensitivity Section in the main paper, where we do see some settings not tested in our main experiments (which we reported metric results for) showing better results than the default ones we used for our experiments).


\begin{table}[!htb]
\begin{center}
\caption{Running time (seconds) per epoch for different datasets. Pytorch is used for modeling and training, and 8 CPUs are employed.}
\label{table:running time}
 \small
\begin{tabular}{  l  | l  }
 \hline
 Dataset & Run time \\\hline
Solar & $600$ \\
Electricity (small) & $770$ \\
Electricity (large) & $1175$ \\
Traffic & $597$ \\
Taxi & $330$ \\
Wiki (small) & $295$ \\
Wiki (large) & $441$ \\
\end{tabular}
\end{center}
\end{table}

\subsection{Model complexity and run times}
\label{sub:computational_complexity}

Denote the input dimension as $n$, $d_j$ the number of hidden units at $j$-th layer of the encoder, $b_{\text{c}}$, $b_{\text{i}}$, and $b_{\text{o}}$ the number of memory cells, input and output units of LSTM, respectively. Then the complexity of the model is
$$
O( ( n d_1 + \sum_j d_j d_{j+1} + b_{\text{i}} d_l + b_{\text{i}} b_{\text{c}} +  b_{\text{c}} b_{\text{o}} + b^2_{\text{c}}  ) L ),
$$
where $L$ is the number of LSTM time steps. If we denote $d$ and $b$ the total of encoder and LSTMs hidden units, respectively, then the model complexity can be simplified to $O((nd + bd) L )$, which is linear with the input dimension assuming that $b, d, L \ll n$. This number is significantly smaller than $n^2$ number of parameters used to model the data distribution directly in the input domain.  

Table \ref{table:running time} shows the running time per epoch of the model with different datasets. For these results we trained the model using Pytorch with $8$ CPUs. We expect the running time to substantially decrease if GPUs are employed. As we can see from the table, although the large wiki dataset contains a significantly larger number of series ($115084$ time series) compared to small wiki (only $2000$ time series), the running time per epoch of the former is just marginally higher than that of the latter. This strongly indicates the scalability of the proposed TLAE. 

Note that a similar relationship holds even if other temporal models are used in the latent space, (e.g., the transformer with self-attention).  The key point is the temporal model introduces the most complexity relative to the number of series and history, and generally requires more complexity to model the complex temporal patterns.  Therefore with the proposed approach, TLAE enables applying the complex temporal model in the lower dimensional latent space instead of the high dimensional input space, thereby reducing the complexity and size of the temporal model (in addition to denoising benefits of modeling in a global aggregate space).  This is a key advantage compared to models applied in the input space, and also makes the proposed approach complementary to other multivariate prediction models, as the TLAE approach can also use these in the latent space in place of LSTM.

\subsection{Benefit of nonlinear factorization - Ablation Study}
\label{sub:ablation}

\begin{table}[htb!]
\centering
\caption{Comparison of linear vs nonlinear encoders with WAPE/MAPE/SMAPE metrics. Lower scores indicate better performance. The best score for each encoder architecture is in bold.}
\label{table:different encoder}
\scriptsize
\begin{tabular}{ l  l  l  }
\hline
 Encoder type & Hidden dim.  & WAPE / MAPE / SMAPE \\
 \hline
Linear FFN & [64, 16] & 0.136 / 0.157 / 0.130 \\
FFN with ReLU & [64, 16] &  \bf 0.107 / 0.120 / 0.095 \\ \hline
Linear FFN & [64, 32] &  0.117 / 0.134 / 0.109 \\
FFN with ReLU & [64, 32] & \bf 0.106 / 0.120 / 0.093 \\
\end{tabular}
\end{table}

This section demonstrates the importance of nonlinear embedding for multivariate time series prediction - a property that is lacking in other global factorization approaches such as TRMF and DeepGLO. In particular, we implement TLAE with different choices of linear and nonlinear encoders and conduct experiments on the traffic dataset. Table \ref{table:different encoder} shows the prediction performance with respect to WAPE/MAPE/SMAPE metrics. One can see that with the same number of hidden dimensions the prediction accuracy is significantly improved on the encoder with ReLU activation function.

Note that metric scores are somewhat improved compared to our main results in the main paper.  This is because we performed this ablation study later on after having run the original set of experiments.  In this case we ran for a larger number of batches using GPUs, beyond the small set of batches we tried in the main experiment.  The point is for the same training setup and architecture, we see significantly better results with nonlinear encoders vs. simple linear encoding, validating the hypothesis that nonlinear embedding can lead to better global modeling of the multivariate time series.


\subsection{Additional comparisons, metrics, and result details}
\label{sub:additional_comparisons}

We provide additional tables \ref{table:CRPS} and \ref{table:MSE} to report probabilistic CRPS and deterministic MSE scores with the std. deviations over multiple runs as well for more thorough comparison - as these would not fit in the main paper. Details of the evaluated metrics are provided in the "Evaluation metrics" Section (later in the supplementary materials).
With MSE metrics, the prediction is calculated as $\hat{\yb}_t = f(\hat{\xb}_t)$ where $\hat{\xb}_t$ is the latent sample mean of the Gaussian distribution (\ref{eq:xi distribution}). As one can see from the tables, when comparing with GP methods \cite{salinas2018copula}, our model achieves better CRPS performance on solar, traffic, and taxi data and comparable performance on electricity and wiki data. Furthermore, TLAE outperforms GPs on MSE metrics for almost all datasets. We make an additional remark that due to the high variability in the taxi data (Figure \ref{fig:taxi series prediction}), GP models admit very high prediction variance while it is much smaller with TLAE. Similarly, with regard to MSE score, GP model produced unreasonably high variance over different runs (much higher than its mean) while TLAE produces consistent prediction over various runs.

\begin{table*}[h!]
\centering
\caption{Comparison of different algorithms with CRPS-Sum metrics. Results are from Table 2 of \cite{salinas2018copula} with our results (under same setup) at the end (TLAE). Lower scores indicate better performance.  Mean and std. dev. are calculated from $3$ separate runs with random initialization. VAR and GARCH are traditional statistical multivariate methods \cite{lutkepohl2005new,bauwens2006multivariate}; Vec-LSTM methods use a single global LSTM that takes and predicts all series at once, with different output Gaussian distribution approaches; and GP methods are DNN gaussian process ones proposed in \cite{salinas2018copula} with GP-Copula the main proposed copula method - see details in \cite{salinas2018copula}. A '-' indicates a method failed (e.g., required too much memory as not scalable enough for data size).}
\label{table:CRPS-sum}
\scriptsize
\begin{tabular}{  l  l  l  l l l }
& CRPS-Sum & & & &\\
\hline
Estimator & Solar & Electricity-Small & Traffic & Taxi & Wiki-Small \\
\hline
VAR & $0.524\pm 0.001$ & $0.031 \pm 0.000$ & $0.144 \pm 0.000$ & $0.292 \pm 0.000$ & $3.400 \pm 0.003$ \\
GARCH & $0.869 \pm 0.000$ & $0.278\pm 0.000$ & $0.368\pm 0.000$ & - & - \\
Vec-LSTM-ind & $0.470 \pm 0.039$ & $0.731 \pm 0.007$ & $0.110 \pm 0.020$ & $0.429 \pm 0.000$ & $0.801\pm 0.029$ \\
Vec-LSTM-ind-scaling & $0.391 \pm 0.017$ & $0.025 \pm 0.001$ & $0.087 \pm 0.041$ & $0.506 \pm 0.005$ & $0.133 \pm 0.002$ \\
Vec-LSTM-fullrank & $0.956 \pm 0.000$ & $0.999 \pm 0.000$ & - & - & - \\
Vec-LSTM-fullrank-scaling & $0.920 \pm 0.035$ & $0.747 \pm 0.000$ & - & - & - \\
Vec-LSTM-lowrank-Copula & $0.319 \pm 0.011$ & $0.064 \pm 0.008$ & $0.103 \pm 0.006$ & $0.326 \pm 0.007$ & $0.241 \pm 0.033$ \\
GP & $0.828 \pm 0.010$ & $0.947 \pm 0.016$ & $2.198 \pm 0.774$ & $0.425 \pm 0.199$ & $0.933 \pm 0.003$ \\
GP-scaling & $0.368 \pm 0.012$ & ${\bf 0.022 \pm 0.000}$ & $0.079 \pm 0.000$ & $0.183 \pm 0.395$ & $1.483 \pm 1.034$  \\
GP-Copula & $0.337 \pm 0.024$ & $ { 0.024\pm0.002}$ & $0.078\pm 0.002$ & $0.208\pm 0.183$ & ${\bf 0.086\pm 0.004}$ \\
{\bf TLAE} & $ \bf 0.124 \pm 0.057 $ & $0.040 \pm 0.003$ & ${\bf 0.069 \pm 0.002 }$ & ${\bf 0.130 \pm 0.010}$ & $0.241 \pm 0.001$ 
\end{tabular}
\end{table*}

\begin{table*}[h!]
\begin{center}
\caption{Comparison of different algorithms with CRPS metrics. Results are from Table 2 of \cite{salinas2018copula} with our results (under same setup) at the end (TLAE). Lower scores indicate better performance.  Mean and std. dev. are calculated from $3$ separate runs with random initialization. VAR and GARCH are traditional statistical multivariate methods \cite{lutkepohl2005new,bauwens2006multivariate}; Vec-LSTM methods use a single global LSTM that takes and predicts all series at once, with different output Gaussian distribution approaches; and GP methods are DNN gaussian process ones proposed in \cite{salinas2018copula} with GP-Copula the main proposed copula method - see details in \cite{salinas2018copula}. A '-' indicates a method failed (e.g., required too much memory as not scalable enough for data size).}
\label{table:CRPS}
\scriptsize
\begin{tabular}{  l  l  l  l  l  l}
& CRPS & & & &  \\
\hline
Estimator & Solar & Electricity-Small & Traffic & Taxi & Wiki-Small \\
\hline
VAR & $0.595 \pm 0.000$ & $0.060 \pm 0.000$ & $0.222 \pm 0.000$ & $0.410 \pm 0.000$ & $4.101 \pm 0.002$ \\
GARCH & $0.928 \pm 0.000$ & $0.291 \pm 0.000$ & $0.426 \pm 0.000$ & - & - \\
LSTM-ind & $0.480 \pm 0.031$ & $0.765 \pm 0.005$ & $0.234 \pm 0.007$ & $0.495 \pm 0.002$ & $0.800\pm 0.028$ \\
LSTM-ind-scaling & $0.434 \pm 0.012$ & $0.059 \pm 0.001$ & $0.168 \pm 0.037$ & $0.586 \pm 0.004$ & $0.379 \pm 0.004$ \\
LSTM-fullrank & $0.939 \pm 0.001$ & $0.997 \pm 0.000$ & - & - & - \\
Vec-LSTM-fullrank-scaling & $1.003 \pm 0.021$ & $0.749 \pm 0.020$ & - & - & - \\
LSTM-lowrank-Copula & $0.384 \pm 0.010$ & $0.084 \pm 0.006$ & $0.165 \pm 0.004$ & $0.416 \pm 0.004$ & $0.247 \pm 0.001$ \\
GP & $0.834 \pm 0.002$ & $0.900 \pm 0.023$ & $1.255 \pm 0.562$ & $0.475 \pm 0.177$ & $0.870 \pm 0.011$ \\
GP-scaling & $0.415 \pm 0.009$ & ${\bf 0.053 \pm 0.000}$ & $0.140 \pm 0.002$ & $0.346 \pm 0.348$ & $1.549 \pm 1.017$  \\
GP-Copula & $0.371 \pm 0.022$ & $0.056 \pm 0.002$ & $0.133 \pm 0.001$ & $0.360 \pm 0.201$ & $\bf 0.236 \pm 0.000$ \\
{\bf TLAE} & $ \bf 0.335 \pm 0.044 $ & $0.058 \pm 0.003$ & ${\bf 0.097 \pm 0.002 }$ & ${\bf 0.369 \pm 0.011}$ & $0.298 \pm 0.002$
\end{tabular}
\end{center}
\end{table*}

\begin{table*}[h!]
\begin{center}
\caption{Comparison of different algorithms with MSE metrics. Results are from Table 2 of \cite{salinas2018copula} with our results (under same setup) at the end (TLAE). Lower scores indicate better performance.  Mean and std. dev. are calculated from $3$ separate runs with random initialization. VAR and GARCH are traditional statistical multivariate methods \cite{lutkepohl2005new,bauwens2006multivariate}; Vec-LSTM methods use a single global LSTM that takes and predicts all series at once, with different output Gaussian distribution approaches; and GP methods are DNN gaussian process ones proposed in \cite{salinas2018copula} with GP-Copula the main proposed copula method - see details in \cite{salinas2018copula}. A '-' indicates a method failed (e.g., required too much memory as not scalable enough for data size).}
\label{table:MSE}
\scriptsize
\begin{tabular}{  l  l  l  l  l  l }
& MSE & & & & \\
 \hline
Estimator & Solar & Electricity-Small & Traffic & Taxi & Wiki-Small \\
\hline
VAR & 7.0e3+/-2.5e1 & 1.2e7+/-5.4e3 & 5.1e-3+/-2.9e-6 & - & - \\
GARCH & 3.5e3+/-2.0e1 & 1.2e6+/-2.5e4 & 3.3e-3+/-1.8e-6 & - & - \\
LSTM-ind & 9.9e2+/-2.8e2 & 2.6e7+/-4.6e4 & 6.5e-4+/-1.1e-4 & 5.2e1+/-2.2e-1 & 5.2e7+/-3.8e5 \\
LSTM-ind-scaling & 9.3e2+/-1.9e2 & 2.1e5+/-1.2e4 & 6.3e-4+/-5.6e-5 & 7.3e1+/-1.1e0 & 7.2e7+/-2.1e6 \\
LSTM-fullrank & 3.8e3+/-1.8e1 & 2.7e7+/-2.3e2 & - & - & - \\
Vec-LSTM-fullrank-scaling & 3.8e3+/-6.9e1 & 3.2e7+/-1.1e7 & - & - & - \\
LSTM-lowrank-Copula & 2.9e3+/-1.1e2 & 5.5e6+/-1.2e6 & 1.5e-3+/-2.5e-6 & 5.1e1+/-3.2e-1 & 3.8e7+/-1.5e5 \\
GP & 3.7e3+/-5.7e1 & 2.7e7+/-2.0e3 & 5.1e-1+/-2.5e-1 & 5.9e1+/-2.0e1 & 5.4e7+/-2.3e4 \\
GP-scaling & 1.1e3+/-3.3e1 & {\bf 1.8e5+/-1.4e4} & 5.2e-4+/-4.4e-6 & 2.7e1+/-1.0e1 & 5.5e7+/-3.6e7 \\
GP-Copula & 9.8e2+/-5.2e1 & 2.4e5+/-5.5e4 & 6.9e-4+/-2.2e-5 & 3.1e1+/-1.4e0 & 4.0e7+/-1.6e9 \\
{\bf TLAE} & {\bf 6.8e2+/-1.3e2} & { 2.0e5+/-1.6e4} & {\bf 4e-4+/-5.0e-6} & {\bf 2.6e1+/-1.4e0} & {\bf 3.8e7+/-7.2e4}
\end{tabular}
\end{center}
\end{table*}

\subsection{Comparison to univariate methods} 
\label{sec:univariate_comparisons}
In this section, we show experiment results comparing our proposed probabilistic TLAE with univariate baselines and state-of-the-art methods \cite{salinas2019deepar,rangapuram2018dsp,li2019transformer} (and a multivariate one \cite{wang2019deep} that used the same metrics) on traffic and small electricity datasets. Train and test sets are set the same as in \cite{li2019transformer}. Following these methods, we evaluate the prediction performance by $\rho$-quantile loss defined in (\ref{eq:quantile}) in the "Evaluation metrics" Section, with our reported scores as the mean over $3$ separate runs. Note that this $\rho$-quantile loss is evaluating the predictive distribution at only 2 points (2 quantiles) as opposed to CRPS which evaluates fit of the entire distribution (i.e., approximated by multiple quantiles distributed through) and better measures how good the uncertainty / distribution prediction is, but we used the $\rho$-quantile loss here to enable fair comparison with the prior work that all used this metric.  Furthermore, many of these methods were specifically trained and targeted at the specific quantiles (i.e., the output is directly targeted at predicting the quantile using a pinball loss function), whereas our model is targeted at modeling the entire distribution (not focusing on particular quantiles - though note that the deterministic version of TLAE could also be targeted at specific quantiles).  Nevertheless, TLAE still gets comparable accuracy on these quantiles to the state-of-the-art, despite fitting the entire distribution simultaneously. 

As one can see from Table \ref{table:quantile}, we obtain comparable scores to Transformer \cite{li2019transformer} - the state-of-the-art method for univariate time series prediction. 
We hypothesize that our result will further be improved if a Transformer is applied in our middle layer. Additionally, we note that while DeepAR \cite{salinas2019deepar}, DeepState \cite{rangapuram2018dsp}, and Transformer \cite{li2019transformer} utilize external features in their training and testing which likely have strong influence on the overall performance, these are not included in our model. We want to again emphasize that our model is complementary to univariate methods. A combination of our global latent forecast features with local features from univariate models as proposed in \cite{sen2019deepglo} will potentially improve the performance further.

\begin{table*}[h!]
\centering
\caption{Univariate time series methods comparison experiment results. Comparison of different algorithms with the reported $\rho$-quantile metrics with $\rho=0.5 / \rho=0.9$. Results were reported from Table 6 of \cite{li2019transformer}  with our results placed at the last columns. Lower scores indicate better performance. The best two scores are in bold.}
\label{table:quantile}
\scriptsize
\begin{tabular}{  l  l  l  l  l  l  l  l l }
\hline
 & ARIMA & ETS & TRMF & DeepAR  & DeepState  & DeepFactor  & Transformer  & TLAE \\
 \hline
Traffic & 0.223/0.137 & 0.236/0.148 & 0.186/- & 0.161 /0.099 & 0.167/0.113 & 0.225/0.159 & {\bf 0.122/0.081} & {\bf 0.117/0.078} \\
Elec. & 0.154/0.102 & 0.101/0.077 & 0.084/- & 0.075 /{\bf 0.040} & 0.083 /0.056 & 0.112/0.059 & {\bf 0.059/0.034} & {\bf 0.072}/0.047 \\
\end{tabular}
\end{table*}



\subsection{Visualization - of losses, predictions, and predictive distributions}
\label{sub:visualization}

{\bf Training losses and evaluated metrics.}

\begin{figure*}
\begin{subfigure}{.5\textwidth}
  \centering
  \includegraphics[width=.99\linewidth]{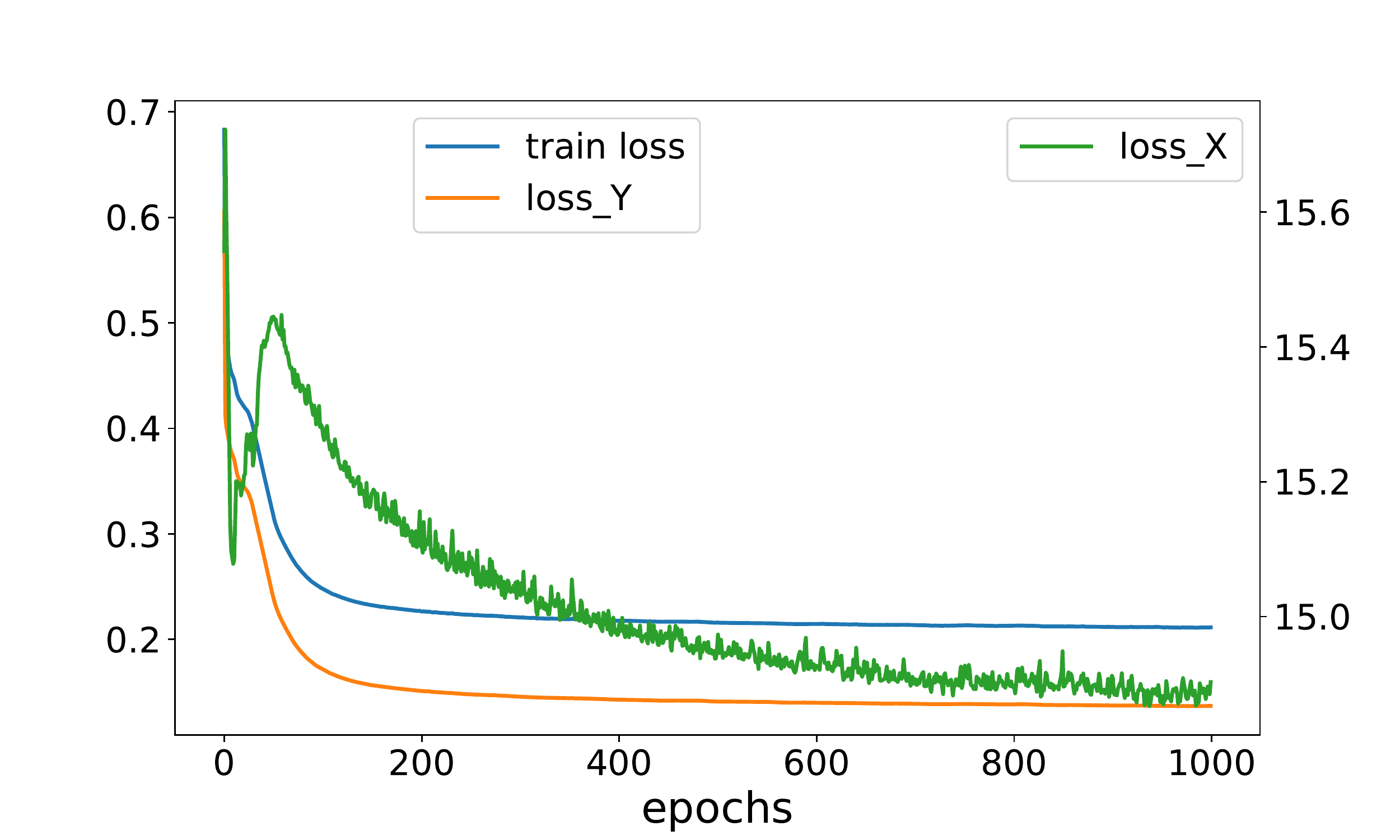}
\end{subfigure}%
\begin{subfigure}{.5\textwidth}
  \centering
  \includegraphics[width=.99\linewidth]{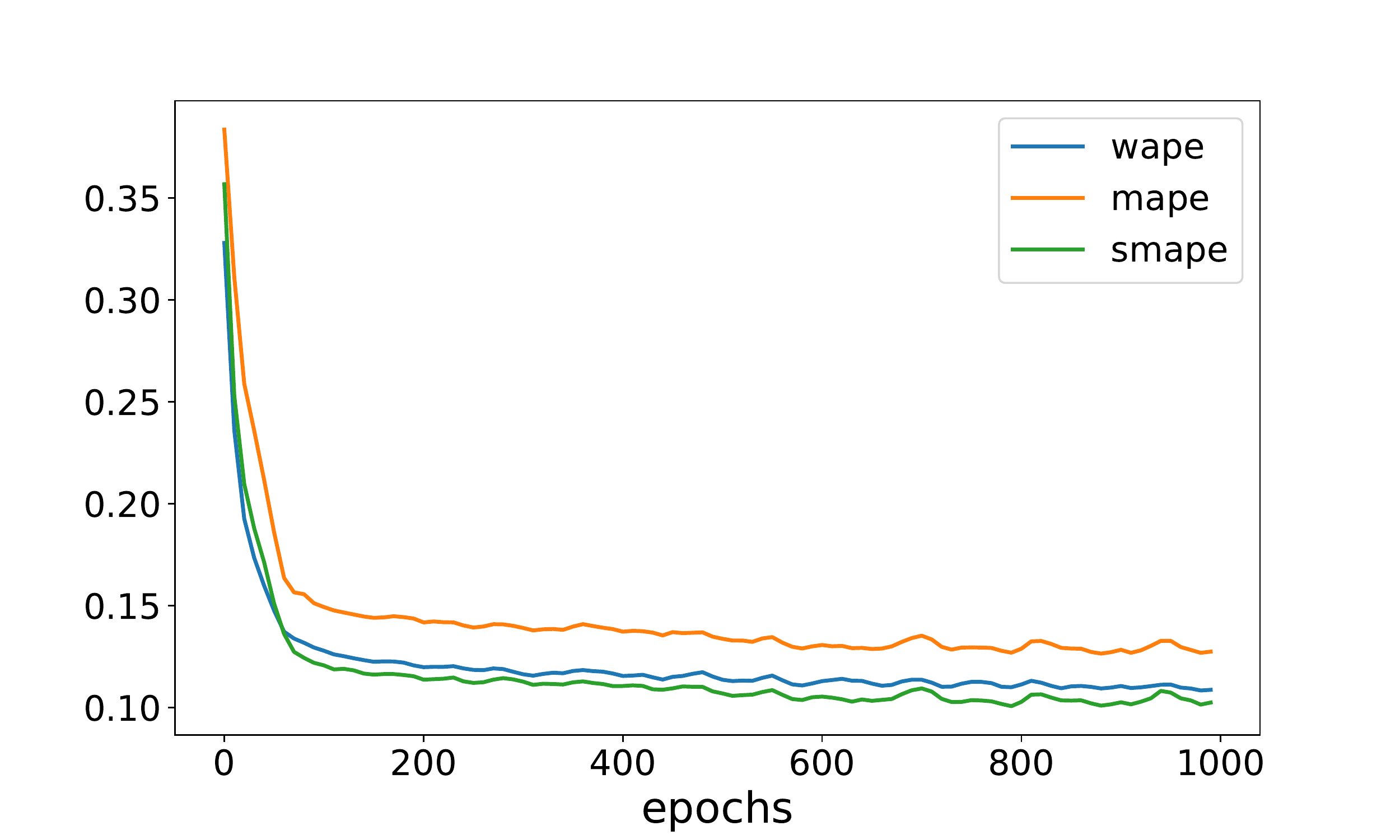}
\end{subfigure}
\caption{Training loss (left) and evaluated metrics (right) versus epochs on the traffic data. On the left, scale of the green curve displaying the loss of $X$ is read from the right while other two is from the left.}
\label{fig:traffic loss and metrics}
\end{figure*}

\begin{figure*}
\begin{subfigure}{.5\textwidth}
  \centering
  \includegraphics[width=.99\linewidth]{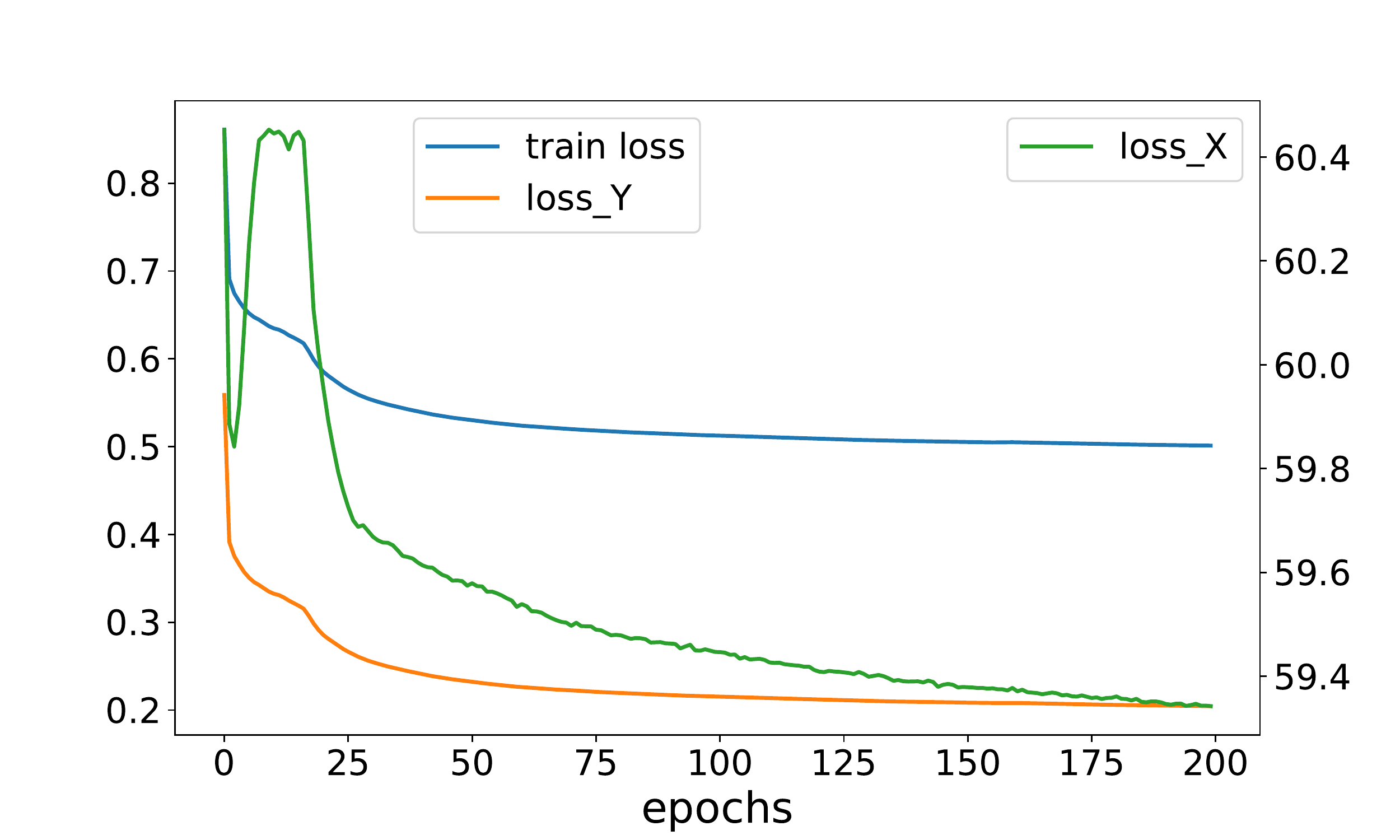}
\end{subfigure}%
\begin{subfigure}{.5\textwidth}
  \centering
  \includegraphics[width=.99\linewidth]{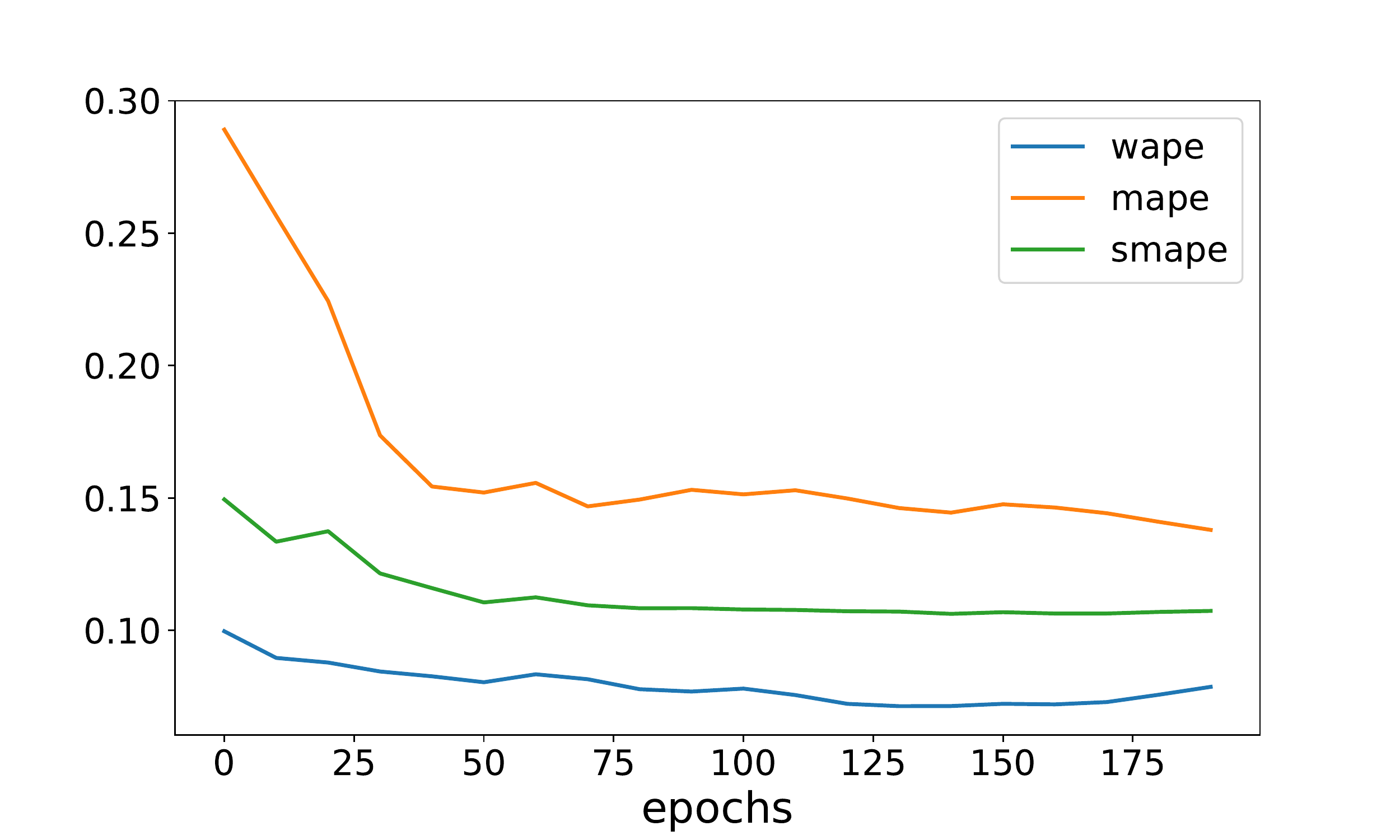}
\end{subfigure}
\caption{Training loss (left) and evaluated metrics (right) versus epochs on the electricity data. On the left, scale of the green curve displaying the loss of $X$ is read from the right while other two is from the left.}
\label{fig:elec loss and metrics}
\end{figure*}

Figure \ref{fig:traffic loss and metrics} shows the training loss and the evaluated test metrics with respect to the training epochs on the traffic data. On the left, the total training loss and the losses on $\Xb$ and $\Yb$ are displayed in which the $\ell_1$ loss is imposed on $\Yb$ while Gaussian negative log likelihood is used for $\Xb$. As one can see, both of these losses decrease with increasing iterations, leading to the convergence of the training loss. Figure \ref{fig:traffic loss and metrics} on the right demonstrates the prediction performance of the test data on the three evaluated metrics. All the metrics make significant improvement in the first $200$ epochs and slowly make progress till the last epoch. We furthermore provide training loss and the test metrics of the electricity dataset in Figure \ref{fig:elec loss and metrics} which presents similar behavior.

{\bf Latent and time series prediction.} 

\begin{figure*}[h!]
\centering
    \includegraphics[width=150mm]{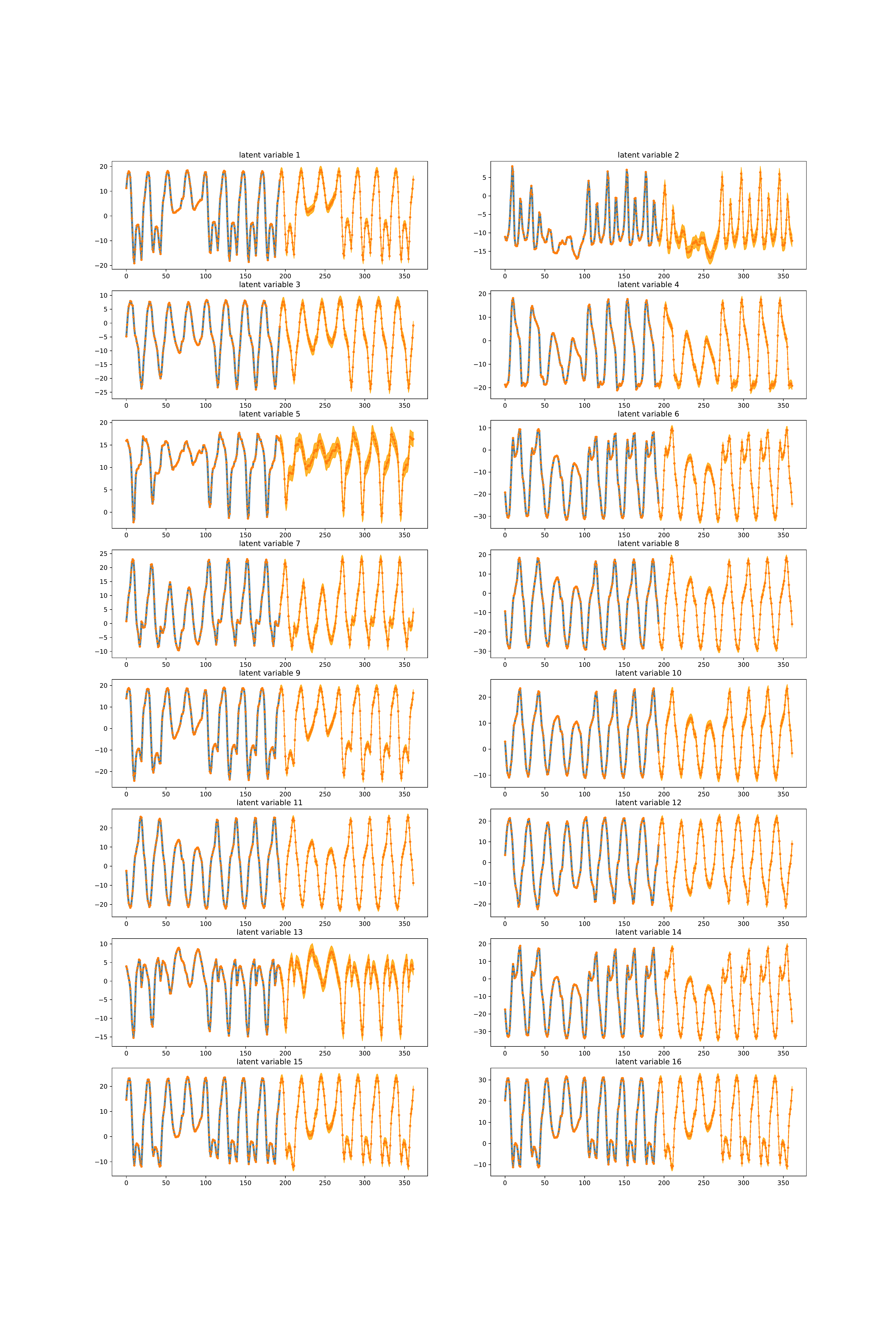}
    \caption{$16$ latent variables and their prediction of the traffic dataset. The first $194$ samples, colored in blue, is the embedding of the multivariate time series $\Yb_B = [\yb_{T-194},..., \yb_T]$. The last $168$ samples, colored in orange, are rolling predictions. The shaded light orange is the $95 \%$ percentile.}
  \label{fig:traffic latent prediction}
\end{figure*}

\begin{figure*}[h!]
\centering
    \includegraphics[width=150mm]{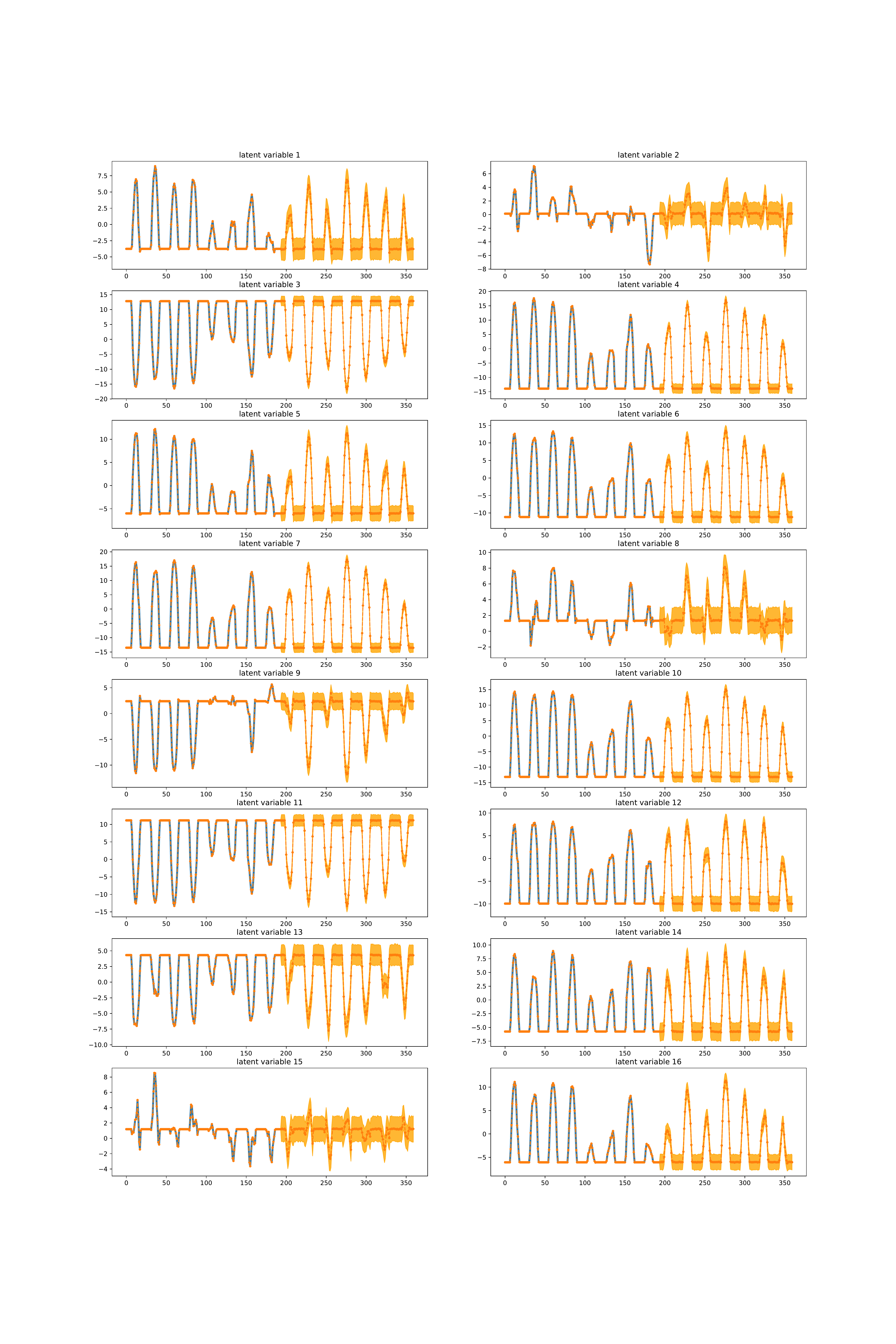}
    \caption{$16$ latent variables and their prediction of the solar dataset. The first $194$ samples, colored in blue, is the embedding of the multivariate time series $\Yb_B = [\yb_{T-194},..., \yb_T]$. The last $168$ samples, colored in orange, are rolling predictions. The shaded light orange is the $95 \%$ percentile.}
  \label{fig:solar latent prediction}
\end{figure*}

\begin{figure*}[h!]
\centering
    \includegraphics[width=150mm]{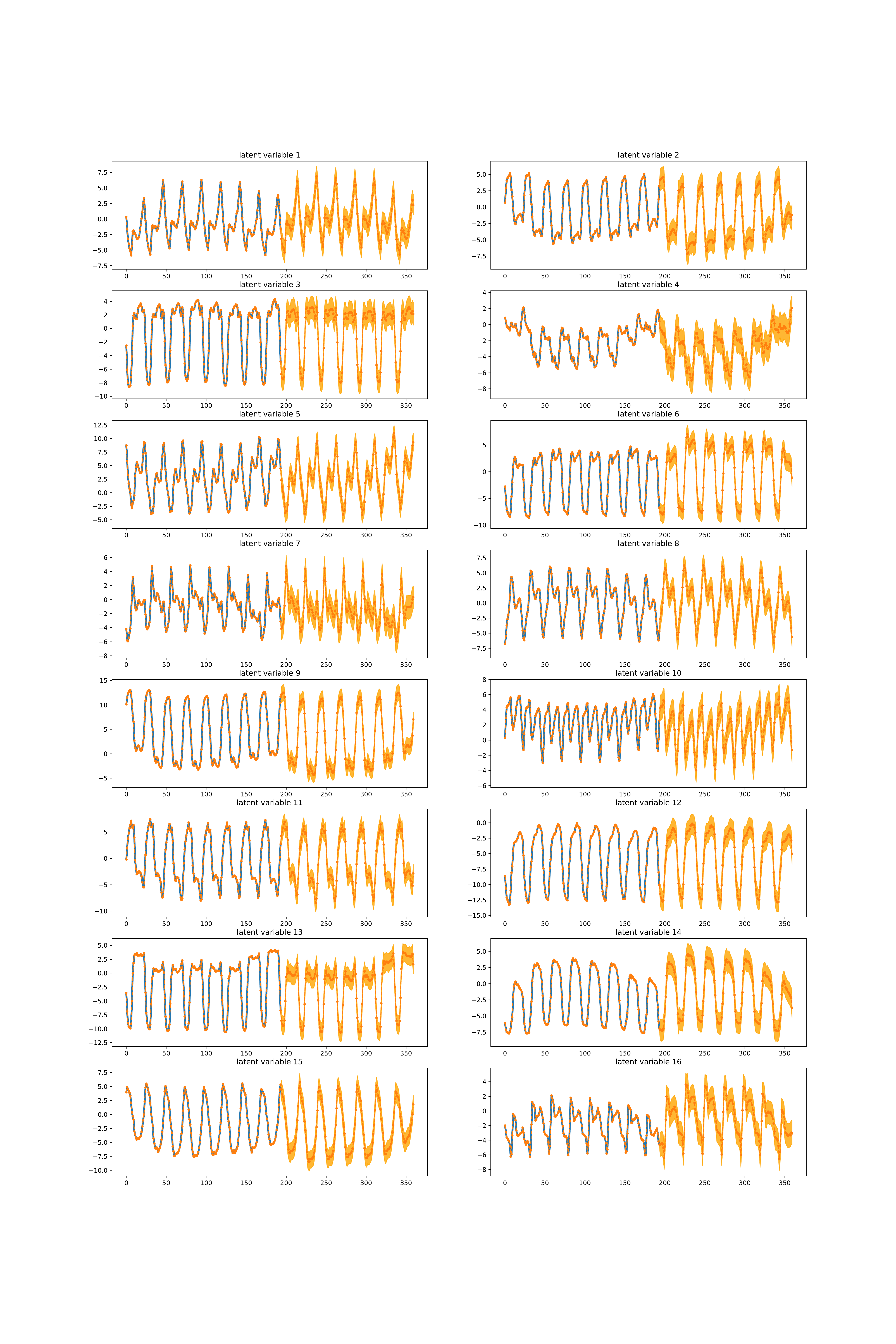}
    \caption{$16$ latent variables and their prediction of the electricity dataset. The first $194$ samples, colored in blue, is the embedding of the multivariate time series $\Yb_B = [\yb_{T-194},..., \yb_T]$. The last $168$ samples, colored in orange, are rolling predictions. The shaded light orange is the $90 \%$ prediction interval ($5 \%$ and $95 \%$ percentiles).}
  \label{fig:elec latent prediction}
\end{figure*}

\begin{figure*}[h!]
\centering
    \includegraphics[width=150mm]{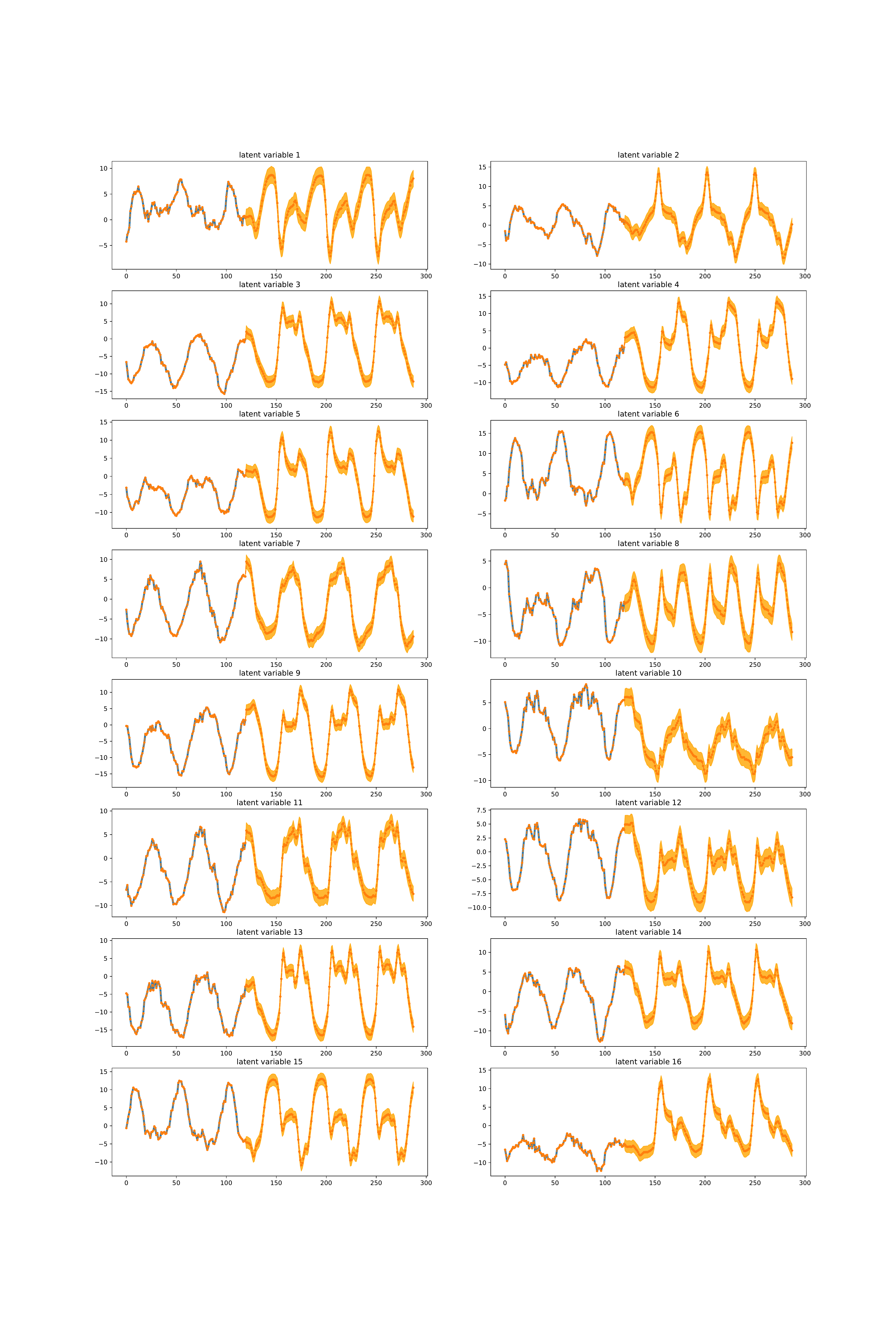}
    \caption{$16$ latent variables and their prediction of the taxi dataset. The first $120$ samples, colored in blue, is the embedding of the multivariate time series $\Yb_B = [\yb_{T-120},..., \yb_T]$. The last $168$ samples, colored in orange, are rolling predictions. The shaded light orange is the $90 \%$ prediction interval.}
  \label{fig:taxi latent prediction}
\end{figure*}

\begin{figure*}[h!]
\centering
    \includegraphics[width=150mm]{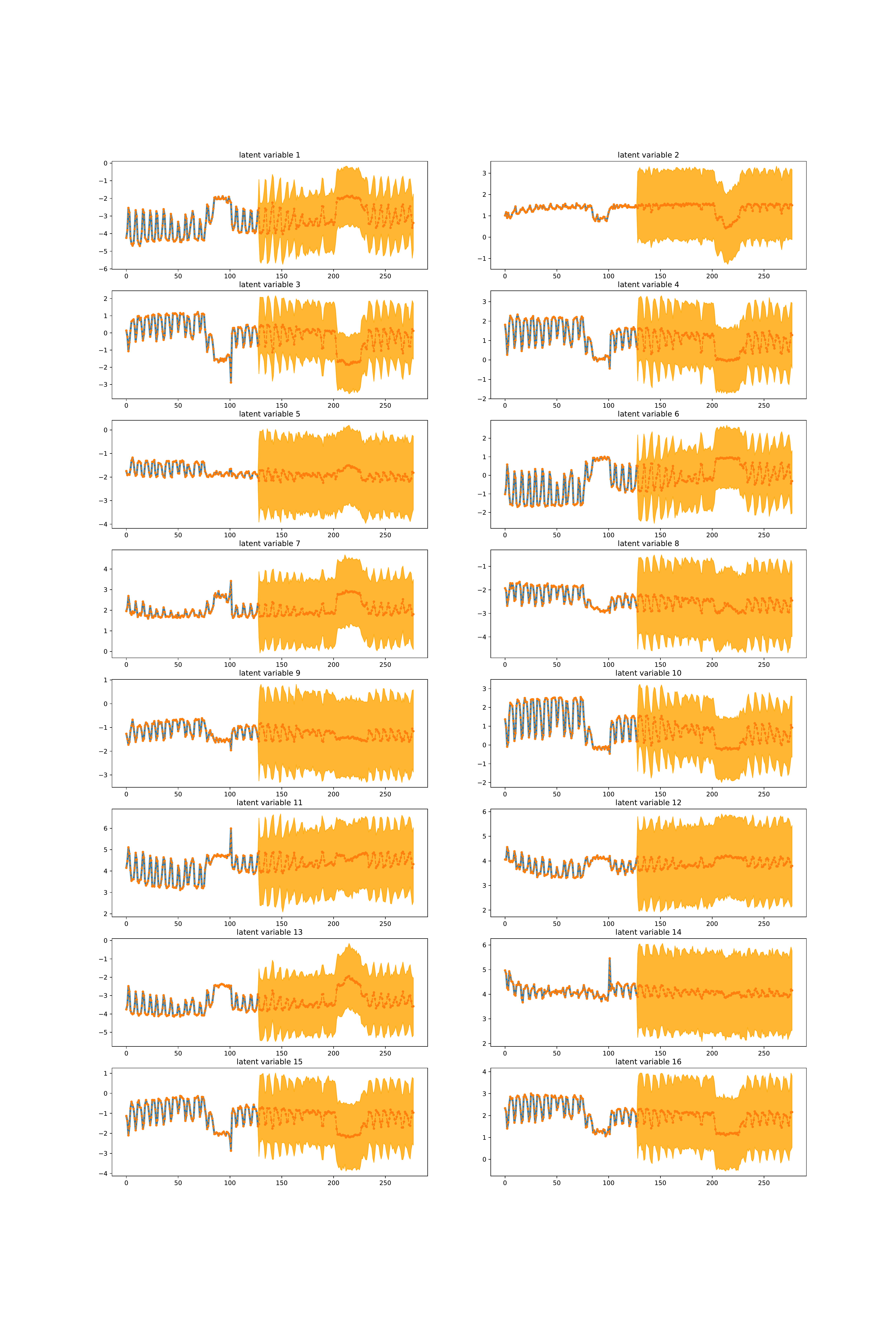}
    \caption{$16$ latent variables and their prediction of the wiki dataset. The first $128$ samples, colored in blue, is the embedding of the multivariate time series $\Yb_B = [\yb_{T-128},..., \yb_T]$. The last $150$ samples, colored in orange, are rolling predictions. The shaded light orange is the $90 \%$ prediction interval.}
  \label{fig:wiki latent prediction}
\end{figure*}

We display the dynamics of trained latent variables and the model's predictions on all the datasets in Figures \ref{fig:traffic latent prediction}, \ref{fig:solar latent prediction}, \ref{fig:elec latent prediction}, \ref{fig:taxi latent prediction}, and \ref{fig:wiki latent prediction}. In the plots of latent variables, embedding variables are shown in blue and the mean predictions in orange. The light shaded area in orange is $90 \%$ prediction interval. The calculation of these variables is as follow: we calculate $L$ latent samples
$$
[\xb_{T-L}, ..., \xb_T] = g_{\bm \phi}(\Yb),
$$
which are mapped from the batch time series data $\Yb = [\yb_{T-L}, ..., \yb_T]$ with $T$ being the total number of training samples. We generate $100$ prediction samples $\hat{\xb}^{(s)}_{T+1}$ from the Gaussian distribution
\begin{equation}
    \label{eq:sample xb at T+1}
    \hat{\xb}^{(s)}_{T+1} \sim \cN( \xb_{T+1} ; h_{\Wb}  (\xb_{T}, ..., \xb_{T-L}), \Ib ), \quad s = 1,..., 100,
\end{equation}
from there, the $95 \%$ percentile, the $5 \%$ percentile, and the prediction mean $\bar{\xb}_{T+1}$ are computed. The next $100$ prediction samples $\hat{\xb}^{(s)}_{T+2}$ is then generated similarly with the predicted input $\bar{\xb}_{T+1}$,
\begin{equation}
    \label{eq:sample xb at T+2}
    \hat{\xb}^{(s)}_{T+2} \sim \cN( \xb_{T+2} ; h_{\Wb}  (\bar{\xb}_{T+1}, \xb_{T}, ..., \xb_{T-L+1}), \Ib ),
\end{equation}
with $s = 1,..., 100$. The iteration is continued for $\tau$ times until the next $\tau$ samples is predicted. Then, the model receives the actual input batch data $\Yb = [\yb_{T-L+\tau}, ..., \yb_{T+\tau}]$ and forecast the next $\tau$ latent samples $\bar{\xb}_{T+\tau+1},..., \bar{\xb}_{T+2\tau}$ with the associated confidence intervals. The procedure is repeated $k$ times to get all $k \tau$ predictions, which are displayed in Figure \ref{fig:traffic latent prediction}. 

As one can observe from these figures, high-level latent variables are capable of capturing global trends across individual time series. Each latent variable possesses its own local properties while they share similar global repeating patterns. 

\begin{figure*}[h!]
\centering
    \includegraphics[width=150mm]{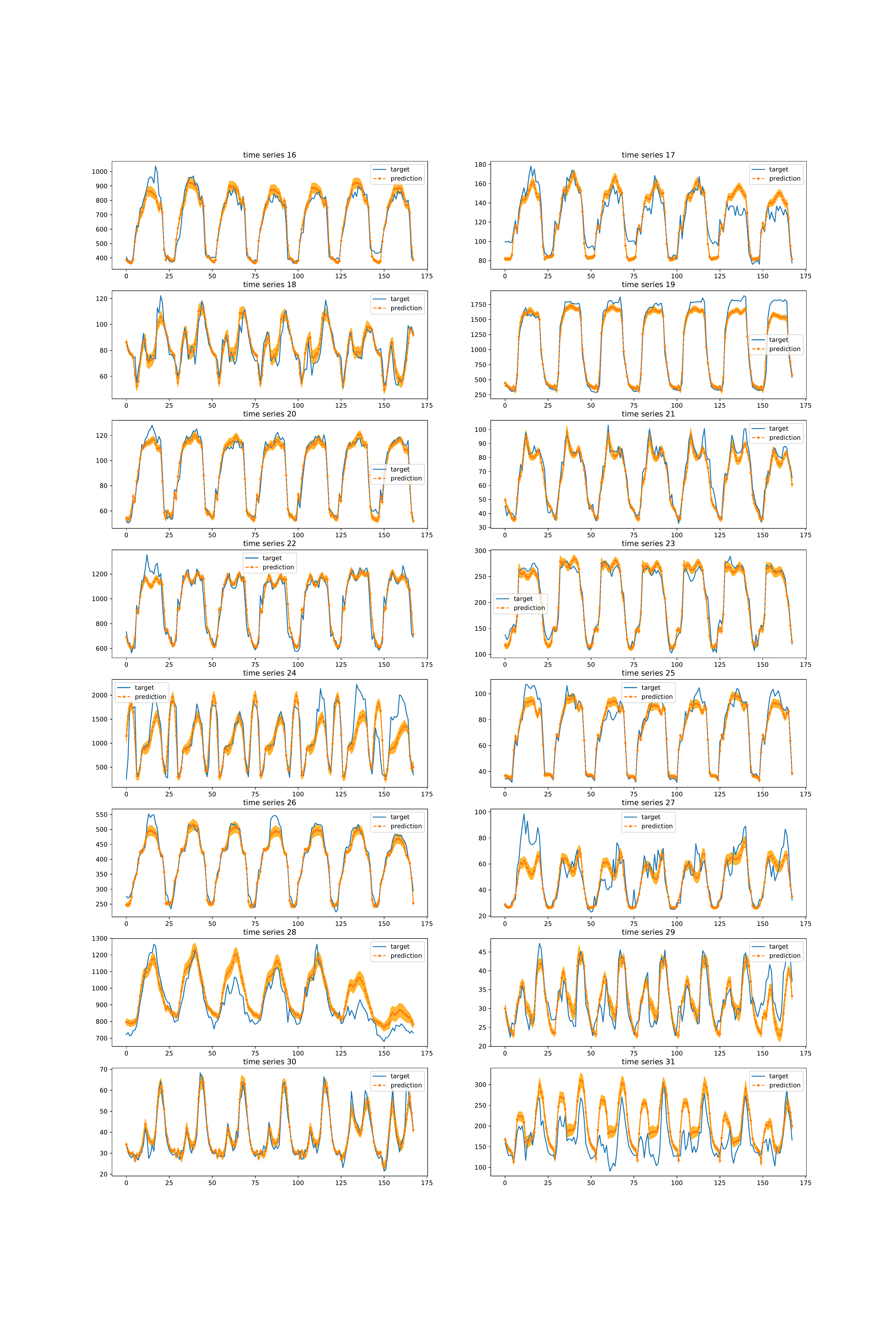}
    \caption{The $168$ prediction samples (orange) of a few time series versus their actual samples (blue) in the electricity dataset. The shaded light orange is the $90 \%$ prediction interval. As one can see, the model not only predict the global patterns of time series well, but also be able to accurately capture local changes in individual series.}
  \label{fig:elec series prediction}
\end{figure*}

\begin{figure*}[h!]
\centering
    \includegraphics[width=150mm]{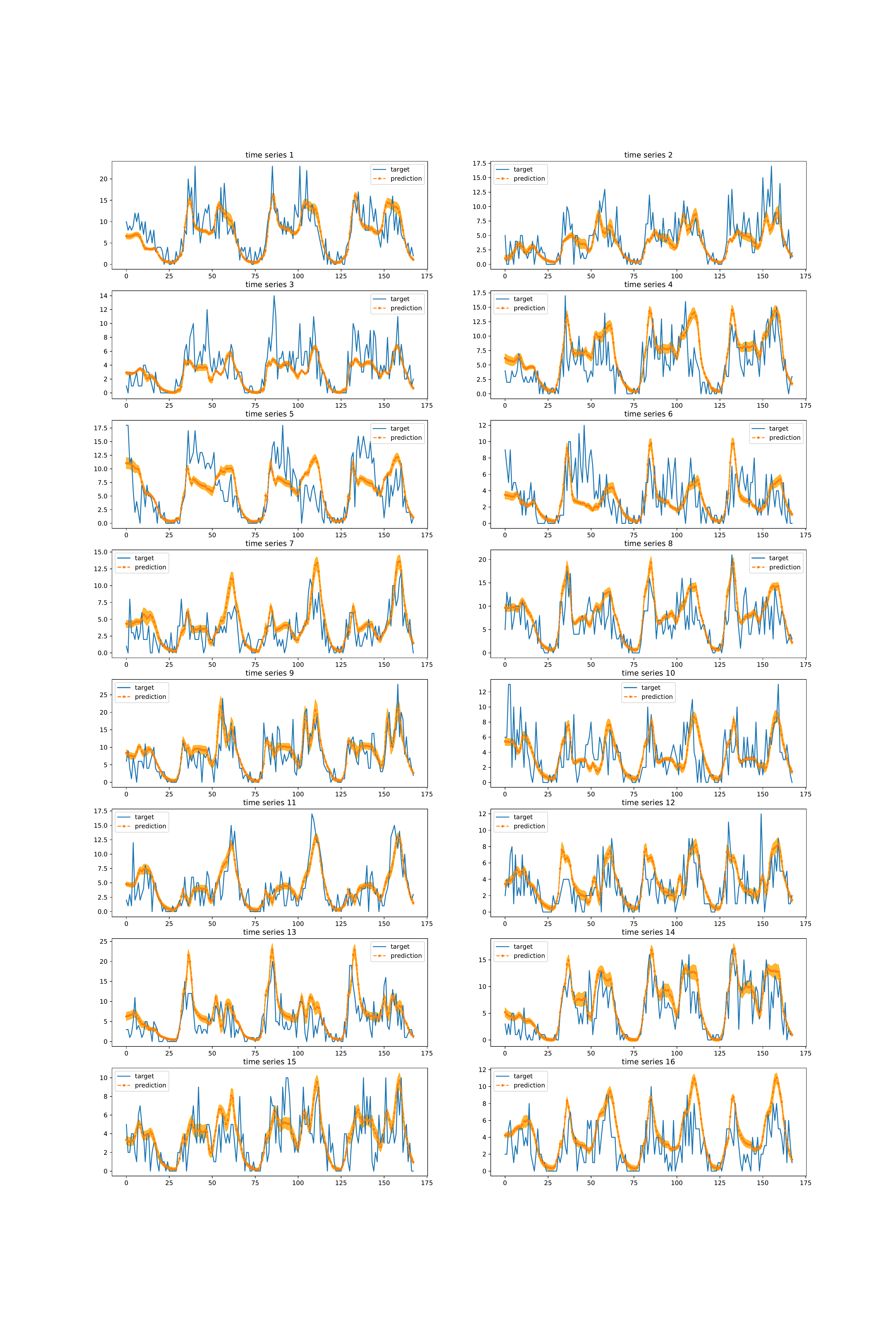}
    \caption{The $168$ prediction samples (orange) of a few time series versus their actual samples (blue) in the taxi dataset. The shaded light orange is the $90 \%$ prediction interval. As one can see, although the data has high variability, the model is able to capture the global pattern.}
  \label{fig:taxi series prediction}
\end{figure*}

\begin{figure*}[h!]
\centering
    \includegraphics[width=150mm]{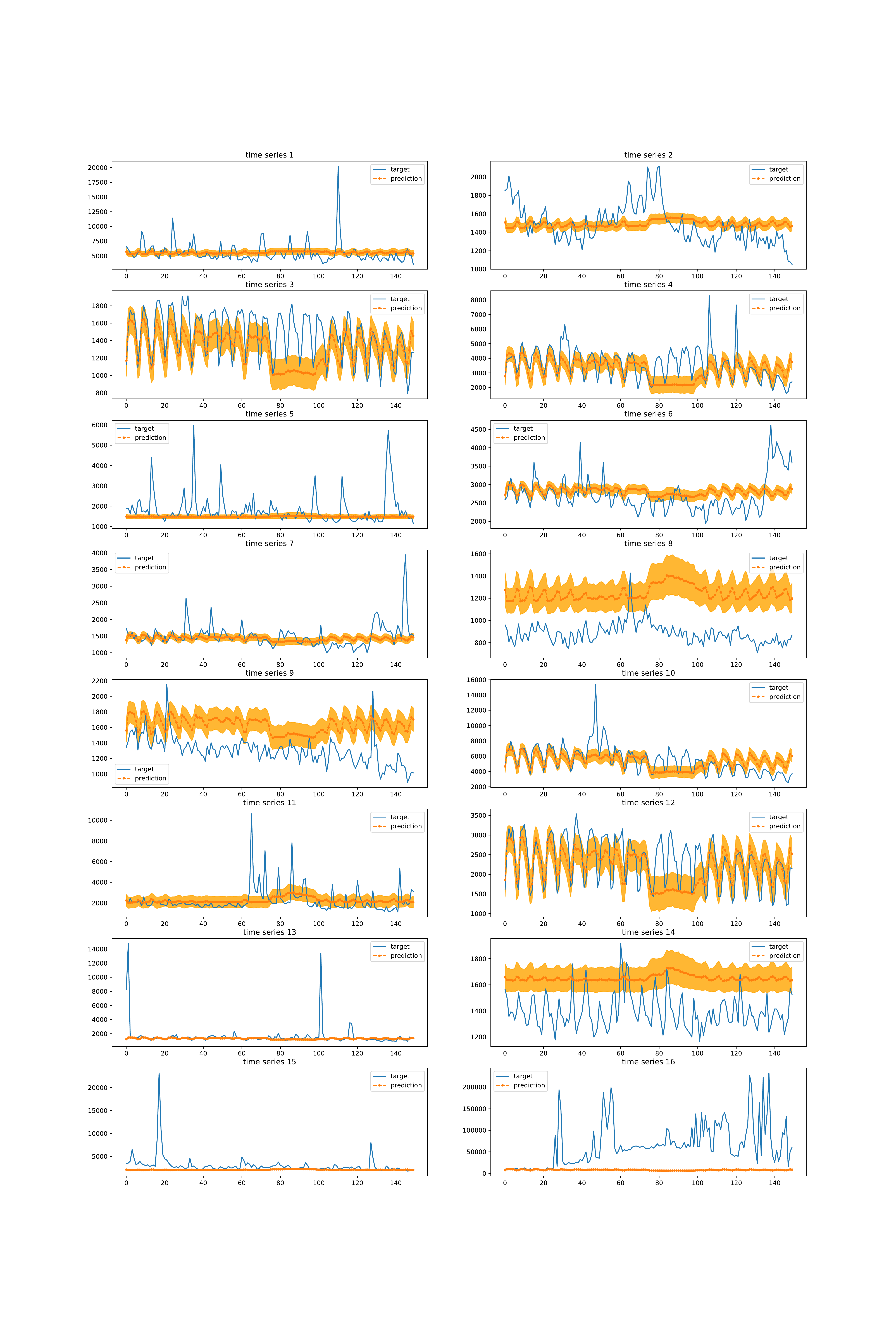}
    \caption{The $150$ prediction samples (orange) of a few time series versus their actual samples (blue) in the wiki dataset. The shaded light orange is the $90 \%$ prediction interval. As one can see, although the data has high variability, the model is able to capture the global pattern of some series and fails on some others.}
  \label{fig:wiki series prediction}
\end{figure*}

In addition, the following figures \ref{fig:traffic series prediction}, \ref{fig:solar series prediction}, \ref{fig:elec series prediction}, \ref{fig:taxi series prediction}, and \ref{fig:wiki series prediction} depict a few actual time series of the underlying datasets and its prediction. The actual series $\Yb = [\yb_{T}, ..., \yb_{T+k\tau}]$ are labeled in blue while the prediction $\bar{\Yb}$ is in orange with its $90 \%$ percentile in light shaded orange color. The output prediction samples are calculated from predicted latent samples in (\ref{eq:sample xb at T+1}) and (\ref{eq:sample xb at T+2}): $\hat{\yb}^{(s)}_{T+i} = f_{\bm \theta} (\hat{\xb}^{(s)}_{T+i})$. As one can see, TLAE can accurately capture the global time series pattern thanks to the predictive capability of the latent representations. In addition, the model is also capable of predicting local variability associated with individual time series.  

Furthermore, we clearly see examples of nonstationary and varying output predictive distributions demonstrating the predictive distribution is not homoskedastic (the output variance changes for different time points and series), despite using a fixed variance in the latent space.   I.e., prediction intervals for individual time series expand and shrink at different time points per series and for different series at the same time (e.g., see Figure \ref{fig:solar series prediction}).  Note this is not surprising as the same idea underpins modern neural network based generative models like VAEs and Generative Adversarial Networks \cite{goodfellow2014generative} - i.e., using a nonlinear decoder to map a simple input distribution to arbitrarily complex output distributions.  The key difference here is this decoding is based on a latent state that is updated over time along with the simple noise distribution, which then gets decoded through the nonlinear network.  In the version of the model we used in experiments, the latent state at a given time point determines how the underlying noise distribution is mapped to the input space distribution (i.e., it must be encoded in the latent state as well).  I.e., even in the current model with feed-forward networks for the encoder and decoder, nonstationary distributions can be modeled through encoding in the latent state and its progression.  However, it's possible to enable potentially even more complex distribution behavior by using temporal models for the encoder or decoder as well - fully testing this is an area of future work.
 

\begin{figure*}[h!]
\centering
    \includegraphics[width=150mm]{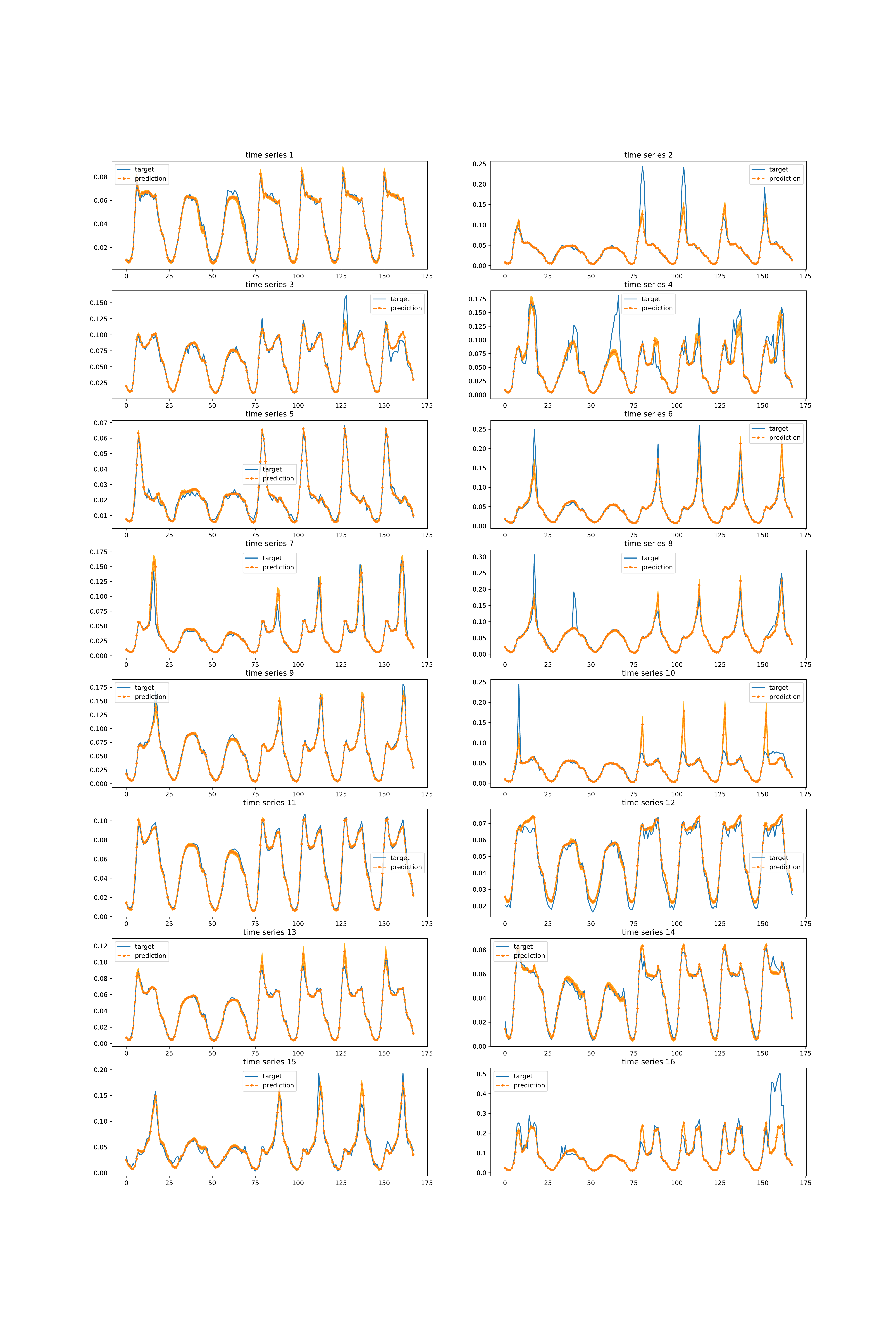}
    \caption{The $168$ prediction samples (orange) of a few time series versus their actual samples (blue) in the traffic dataset. The shaded light orange is the $95 \%$ percentile. As one can see, the model not only predict the global patterns of time series well, but also be able to accurately capture local changes in individual series.}
  \label{fig:traffic series prediction}
\end{figure*}

\begin{figure*}[h!]
\centering
    \includegraphics[width=150mm]{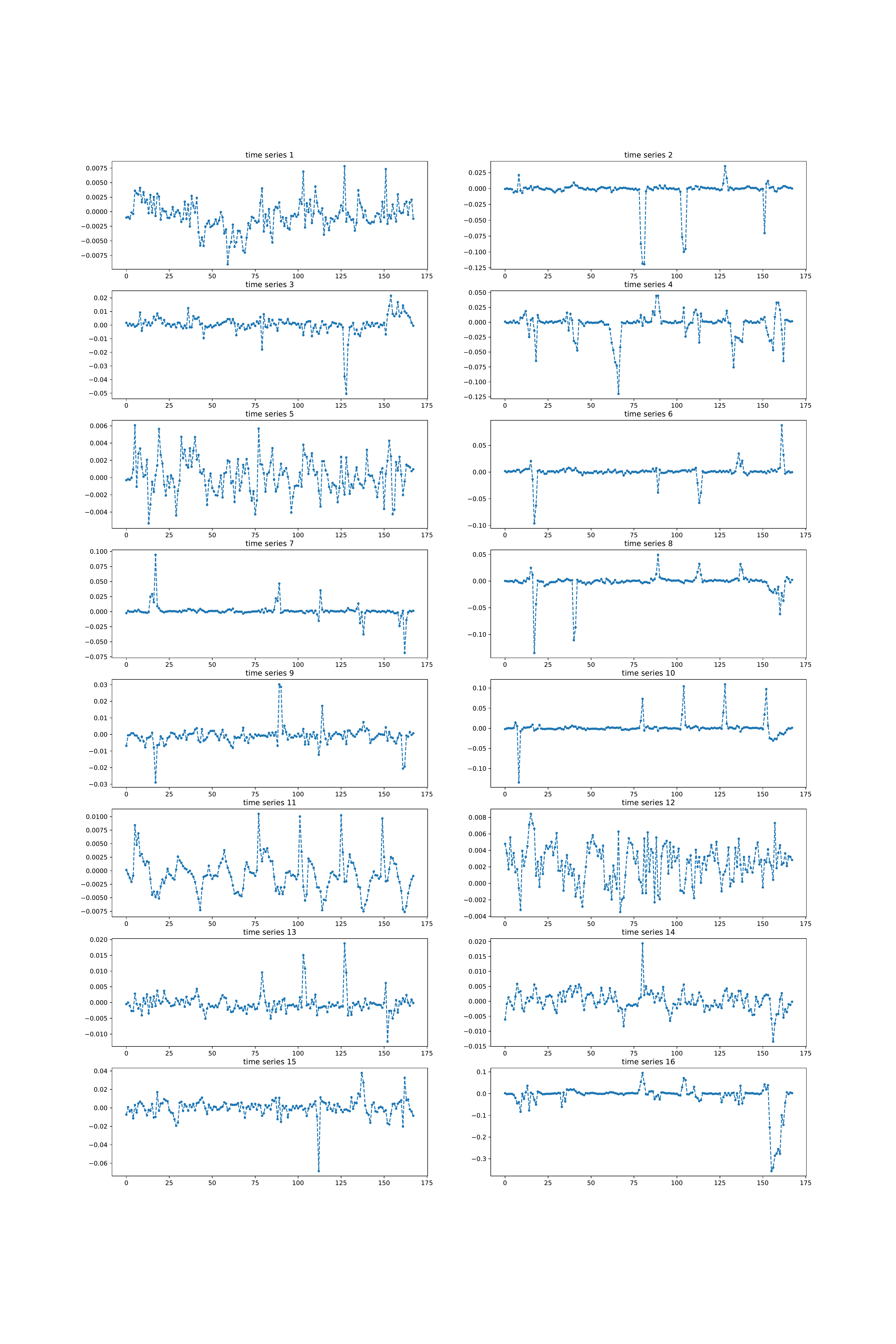}
    \caption{The prediction errors of a few time series in the traffic dataset. As one can see, the prediction error does not exhibit any periodic pattern as shown in the original series in Figure \ref{fig:traffic series prediction}. This indicates that not much information can be further extracted from this noise to improve the prediction.}
  \label{fig:traffic series prediction error}
\end{figure*}

\begin{figure*}[h!]
\centering
    \includegraphics[width=150mm]{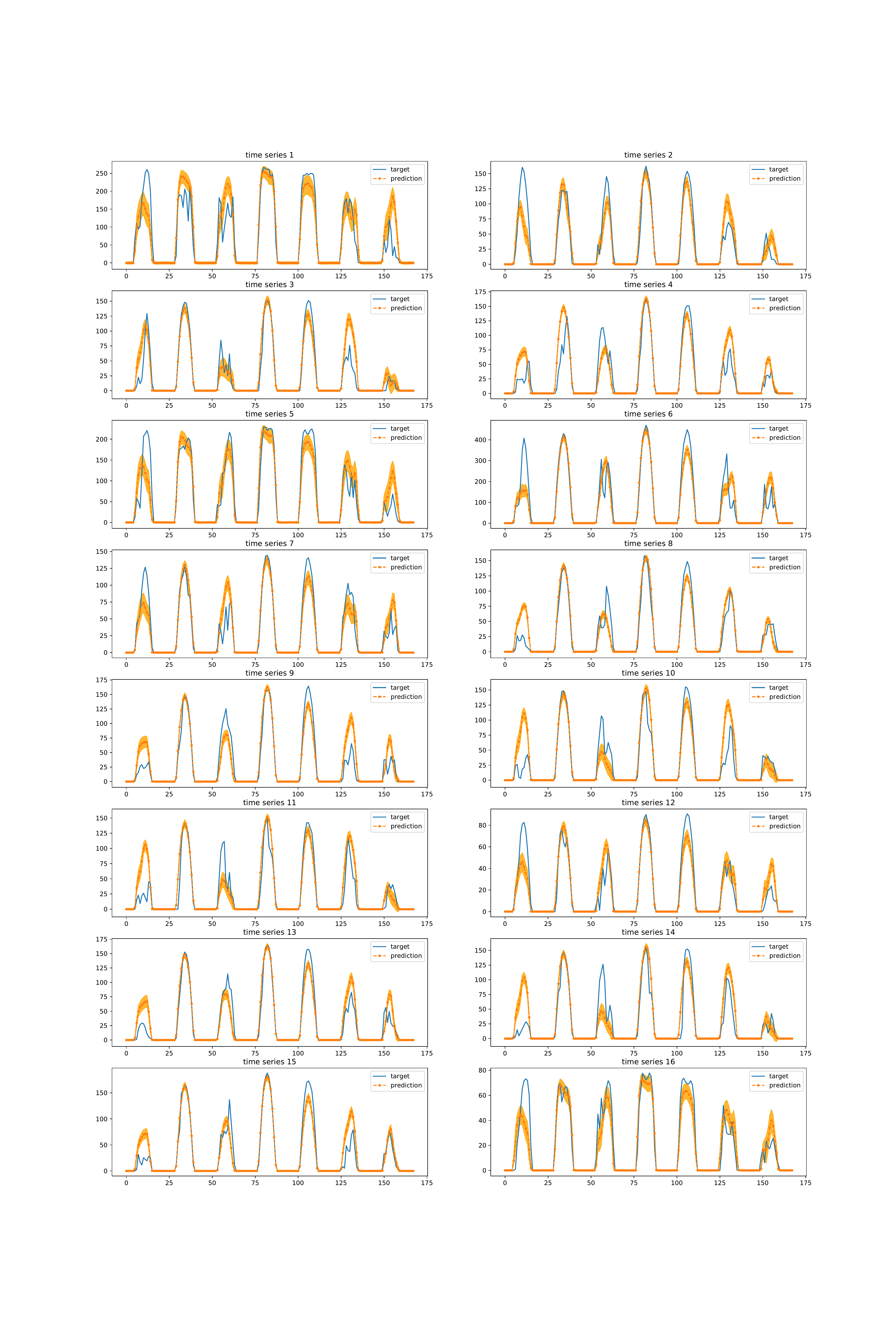}
    \caption{The $168$ prediction samples (orange) of a few time series versus their actual samples (blue) in the solar dataset. The shaded light orange is the $95 \%$ percentile. As one can see, the model not only predict the global patterns of time series well, but also be able to accurately capture local changes in individual series.}
  \label{fig:solar series prediction}
\end{figure*}

{\bf Joint predictive distribution.}

\begin{figure*}[h!]
\centering
    \includegraphics[width=\linewidth]{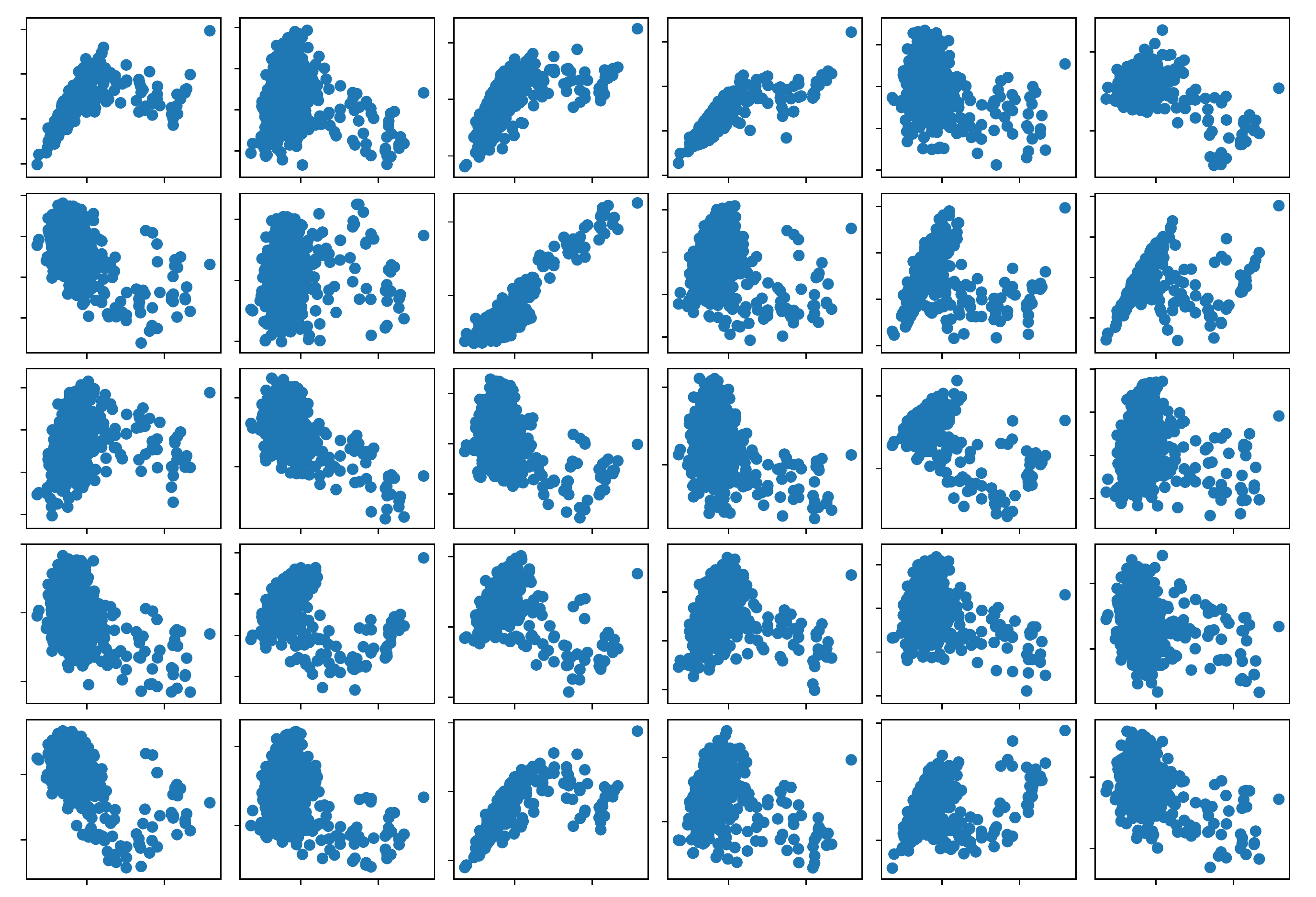}
    \caption{Scatter plots of one series with respect to others on the traffic dataset. Each dot in the subplot indicates a single prediction sample. A few plots attain diagonal structure (indicates strong correlation between two time series) while majority admit multimodal behavior.}
  \label{fig:traffic correlation}
\end{figure*}

Figure \ref{fig:traffic correlation} demonstrates the correlation of one predicted time series with respect to some others on the traffic data. Given $1000$ prediction samples $\hat{\yb}_{T+1}^{(s)}$ decoded from $f_{\bm \theta} (\hat{x}_{T+i}^{(s)})$ where latent $\hat{x}_{T+i}^{(s)}$ is drawn from Gaussian distribution with $s=1,...,1000$, each subplot shows the scatter distribution of the first series $\hat{\yb}_{T+1, 0}^{(s)}$ with another $\hat{\yb}_{T+1, i}^{(s)}$ for $i=1,...,30$, where each dot is a single prediction sample. We observe that a few plots attain diagonal structure (indicating strong correlation between two time series) while the majority exhibit multimodal distributions and nonlinear relationships. These plots suggest that the marginal join distribution between two time series that TLAE captures is much more complex than the joint Gaussian distribution, thanks to the expressive capability of the deep decoder. 

\subsection{Extension to tensor data for incorporating exogenous variables}
\label{sub:tensor_extension}
Although we did not conduct experiments with this setting, we would like to comment on the capability of our model to handle higher dimensional data - i.e., to incorporate additional exogenous features (other than the time series history themselves) in the modeling and forecasting.  So far in our discussion, we assume the multivariate data is two dimensional where one dimension is temporal and the other consists of different time series. In many practical settings, each time series may be associated with additional temporal features (even static features can be modeled as a constant over time temporal feature). For example, for retail demand forecasting, where each series is a combination of a product and a store - weather forecasts for the store region can be one temporal feature, current and planned future price of the product at the given node can be another, events at each location another, seasonal indicators like day of week and hour of day another, etc.  

In the case of global features (those that are the same for all series, like day of week, hour of day, etc.) - these could directly be incorporated as exogenous inputs to the latent space temporal model used (e.g., LSTM) - as these are the same for all series.  However, for additional features that are local to specific series, a more scalable approach as needed.

We can therefore capture the data including exogenous features per time series as a three-dimensional tensor instead of a two dimensional matrix. In this case now $\Yb \in \RR^{n \times p \times T }$ where $p$ is the number of additional features per series.  I.e., each slice of the tensor along the time dimension at time point $t$ is composed of an $n \times p$ matrix $\Yb_t$ of the $n$ time series and each of their $p$ associated features.  Notice that some of these exogenous features may be the same for either some subset of series, all series, all time points, or some combination of these, however, this is not a problem as these can just be included as repeated values in the $3$D tensor. 

In this formulation the encoder still maps the input to the same $d$ dimensional latent space and associated $2$D matrix $\Xb$, so the latent space modeling remains unchanged.  In this case the encoder  $g_{\bm \phi}$ maps $\Yb$ to $d$ dimensional $\Xb$, per time point: $g: \RR^{n \times p} \rightarrow \RR^d$.  There are multiple ways to accomplish this, one trivial way is to add a flattening layer as the first layer of the encoder that maps $\RR^{n \times p} \rightarrow \RR^{np}$, as was done in early work applying deep learning to image classification.  Alternatively, we can borrow the ideas used for image data and use convolutional neural net (CNN) style architectures to capture the cross-series and feature relationships, at lower compute cost and fewer model parameters.  Further, if we want to feed prior time points in as well to influence the embedding for the current time point, we can either feed in a fixed time window of $\Yb$ and perform convolution over the $3$D tensor (i.e., temporal convolution as in TCNs, but now with additional channels to start), or use a combination with other recurrent structures like LSTMs.  The decoder can similarly map back to the original tensor input space, e.g., using deconvolution.  

\subsubsection{Updated losses}
In this case, the loss for predictions in the latent space and the reconstruction loss remains the same.  However for the component of the loss on $\hat{\Yb}$ for the prediction in the input space, $\hat{\yb}_i$ with $i=L,...,b$ (those components that are a function of the past input), it may be desirable to only enforce predicting accurately the future values of the multivariate time series themselves, as opposed to the exogenous features - as these exogenous features might be provided for future values and so need not be forecast.  In this case we can use masking on the second loss term measuring the discrepancy between $\Yb$ and $\hat{\Yb}$ to $0$-out the contribution of the exogenous feature elements for the prediction time points, during training.

\subsubsection{Multistep prediction}
This formulation is fine for predicting the next step, but a common issue with using exogenous features with multi-step prediction is how to get the future values of the exogenous variables, to feed in for predicting subsequent steps.  In some cases these will be known and can directly be plugged in at each next step (for example, day of week, hour of day, holidays, and other seasonal features are known for future time points as well).  In others, forecasts for these might be available or provided separately as well (for example, for weather, sophisticated weather forecast models can project next step weather feature values).  However, if neither is possible, the model itself can be used to forecast the next values of the exogenous features, as these are naturally outputs of the end-to-end model.  I.e., it can be used in the same way as before with no external features - the predicted outputs in the latent space can be used to feed the next step prediction.

Another alternative approach is to change the model to a multi-time-step output model - i.e., train the model to simultaneously predict multiple next time points, given the input.  The disadvantage of this approach is that the model can only be applied then for a specific number of next steps it was trained for, and different models would be needed for different number of next steps.  E.g., if we wanted to predict one more next step than it was trained for, to use this approach we would have to train another model with one additional output.


\subsection{Evaluation metrics}
\label{subsec:evaluation metrics}
Here we describe the evaluation metrics used for both the point (deterministic) and probabilistic forecast settings in more detail.
\subsubsection{Deterministic metrics}
For deterministic settings, we employ well-known metrics that are often used in time series for comparison. In the following, $\Yb \in \RR^{n \times d}$ is denoted as the target time series while $\hat{\Yb} \in \RR^{n \times d}$ is the prediction.
\begin{enumerate}
    \item Weighted Absolute Percentage Error (WAPE) is defined as
    $$
    \frac{\sum_{i=1}^n \sum_{j=1}^d |\hat{y}_{ij} - y_{ij}|}{\sum_{i=1}^n \sum_{j=1}^d |y_{ij}|}.
    $$
    \item Mean Absolute Percentage Error (MAPE) is defined as 
    $$
    \frac{1}{m} \sum_{i=1}^n \sum_{j=1}^d \frac{ |\hat{y}_{ij} - y_{ij}|}{|y_{ij}|} \bm 1_{\{|y_{ij}| > 0 \}},
    $$
    where $m = \sum_{i=1}^n \sum_{j=1}^d \bm 1_{\{|y_{ij}| > 0 \}}$.
    \item Symmetric Mean Absolute Percentage Error (SMAPE) is defined as 
    $$
    \frac{1}{m} \sum_{i=1}^n \sum_{j=1}^d \frac{ 2 |\hat{y}_{ij} - y_{ij}|}{|\hat{y}_{ij} + y_{ij}|} \bm 1_{\{|y_{ij}| > 0 \}},
    $$
    where $m = \sum_{i=1}^n \sum_{j=1}^d \bm 1_{\{|y_{ij}| > 0 \}}$.
    \item Mean Square Error (MSE) is defined as
    $$
    \frac{1}{nd} \sum_{i=1}^n \sum_{j=1}^d (\hat{y}_{ij} - y_{ij})^2.
    $$
\end{enumerate}
We note that these metrics can also be applied when the model is probabilistic. In this case, $\hat{\Yb}$ will be the mean of the predicted distribution.

\subsubsection{Probabilistic metrics}
\label{subsec:prob metrics}
\begin{figure*}[!htbp]
	\centering
	\begin{subfigure}[t]{0.48\textwidth} 
		\includegraphics[width=\textwidth]{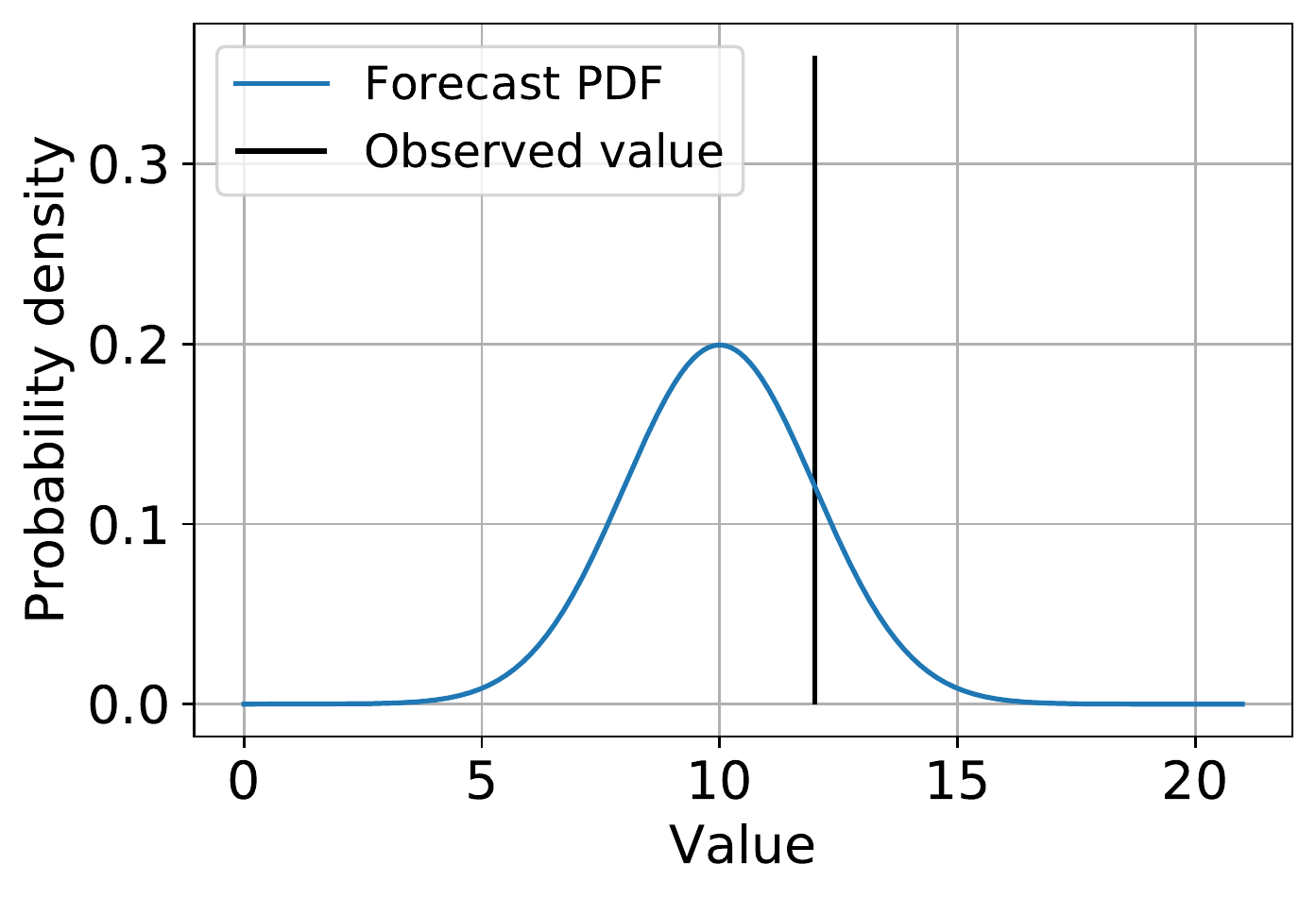}
	\end{subfigure}
	\begin{subfigure}[t]{0.48\textwidth} 
		\includegraphics[width=\textwidth]{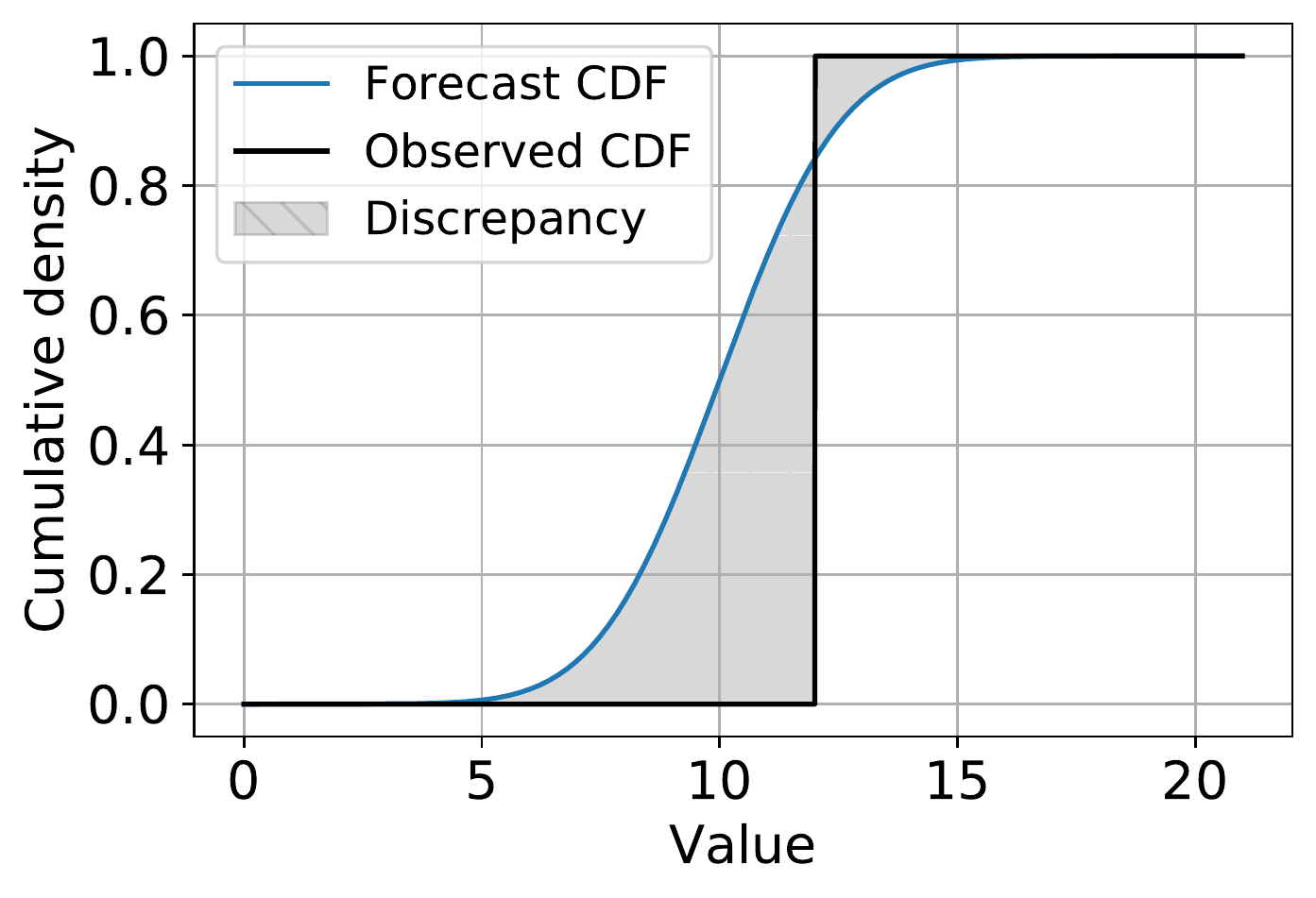}
	\end{subfigure}
	\caption{Illustration of CRPS metric calculation per prediction / observation.  The predictive distribution (PDF) along with the observation is illustrated in (a).  The CDF of the predictive distribution, and the CDF of the observation (Heaviside function) are shown in (b). By comparing the discrepancy between the two (shaded region in b) we can determine how good a fit the predicted distributions are to the observed values, averaged across many different observations.}
	\label{fig:CRPS}
\end{figure*}

To compare with other probabilistic models on time series forecasting, as in \cite{salinas2018copula}, we use the widely-used Continuous Ranked Probability Score (CRPS) \cite{matheson1976scoring,gneiting2007strictly,salinas2018copula,jordan2019evaluating}, which measures the mean discrepancy between a probability distribution (i.e., predictive distribution) $p$ and a deterministic observation $y$.  The CRPS is able to effectively measure the fit of the predictive distribution to the true one and is a \emph{proper scoring rule} - i.e., it is minimized by the true underlying distribution \cite{matheson1976scoring,gneiting2007strictly,jordan2019evaluating}.
Specifically to measure how well the predictive distribution fits across test cases, it computes the integral of the squared difference  between the cumulative density function (CDF) of the predictive distribution and the observed CDF (Heaviside function), as illustrated in Figure \ref{fig:CRPS}, which is a generalization of mean absolute error to the predictive distribution case \cite{gneiting2007strictly}.  CRPS for a sample of observations and forecasts is defined as:
\begin{equation}
\label{eq:CRPS}
    \text{CRPS}(F^f,F^o) = \frac{1}{nd} \sum_{i=1}^n \sum_{j=1}^d \int_{-\infty}^\infty (F^f_{ij}(x) - F^o_{ij}(x))^2 dx
\end{equation}

where $F^f_{ij}(x)$ is the forecast CDF for the $ij$ point (time point and series being predicted) and $F^o_{ij}(x)$ is the CDF of the corresponding observation (represented by the Heaviside function - i.e., $0$ for values less than the observed value and $1$ for values greater than or equal to the observed value).

As we do not have a closed-form for the predictive distribution and CDF, we evaluate the CRPS using a sample generated from the predictive distribution.  Specifically, we borrow the approach and implementation of \cite{salinas2018copula} to compute the CRPS using the alternative pinball loss definition in which the integral is closely approximated by evaluating quantiles of samples from the predictive distribution (see the supplementary material, Section G.1, in \cite{salinas2018copula} for complete details).  
To compute this we sample 1000 points from the predictive distribution for each prediction point, and evaluate the CRPS using 20 quantiles as was done in \cite{salinas2018copula} (note our implementation was directly taken from the implementation shared by the authors used to generate the results reported in \cite{salinas2018copula}  - see \url{https://github.com/mbohlkeschneider/gluon-ts/tree/mv_release} for specific code).  Similarly as in \cite{salinas2018copula} we report the CRPS-sum metric to capture joint effects - which is the CRPS computed on the sum (across series) of the observed target values and the statistics on the sum of the predictive distribution samples.

Another set of metrics that we use in the paper is the $\rho$-quantile loss $R_{\rho}$ with $\rho \in (0,1)$,
\begin{equation}
\label{eq:quantile}
    R(\hat{\Yb}, \Yb) = \frac{2 \sum_{i=1}^n \sum_{j=1}^d D_{\rho}(\hat{y}_{ij}, y_{ij})}{\sum_{i=1}^n \sum_{j=1}^d |y_{ij}|} ,
\end{equation}
    where
    $$
    D_{\rho}(\hat{y}, y) = (\rho - \Ib_{ \{ \hat{y} \leq y \} }) (\hat{y} - y),
    $$
    where $\hat{y}$ is the empirical $\rho$-quantile of the predicted distribution and $\Ib_{ \{ \hat{y} \leq y \}}$ is an indicator function.

\end{document}